\documentclass[lettersize,journal]{IEEEtran}
\usepackage{amsthm}

\usepackage{algorithmic}
\usepackage{algorithm}
\usepackage{array}
\usepackage[caption=false,font=normalsize,labelfont=sf,textfont=sf]{subfig}
\usepackage{textcomp}
\usepackage{stfloats}
\usepackage{url}
\usepackage{verbatim}
\usepackage{graphicx}
\usepackage{cite}
\usepackage{algorithm}
\usepackage{amsmath}
\usepackage{algorithmic}
\usepackage{lettrine}
\usepackage{amssymb}
\usepackage{pifont}

\usepackage{bm}

\usepackage{amsthm} 
\newtheorem{proposition}{Proposition}

\usepackage{xcolor} \usepackage[ colorlinks=false, citebordercolor={0 1 0}, linkbordercolor={0 0 1},  urlbordercolor={0 0 1}, pdfborder={0 0 1} ]{hyperref}

\usepackage{booktabs}
\usepackage{soul}
\usepackage{colortbl}
\sethlcolor{yellow}
\soulregister{\cite}{7}\soulregister{\ref}{7}\soulregister{\eqref}{7}
\soulregister{\emph}{1}\soulregister{\textbf}{1}

\begin{document}

\title{InvarOrd: Self-Supervised Ordinal Knowledge Distillation with Rank-Invariant Alignment for Label-Efficient Cross-Species Hyperspectral Fish Freshness Assessment}

\author{Kazi~Nabiul~Alam,
Omenuko~Joseph~Nwebonyi,
Pooneh~Bagheri~Zadeh,
and~Akbar~Sheikh-Akbari
 
\thanks{The authors are with the School of Built Environment, Engineering and Computing, Leeds Beckett University, Leeds, United Kingdom.}
}

\markboth{IEEE Transactions on AgriFood Electronics}%
{Alam \MakeLowercase{\textit{et al.}}: InvarOrd for Cross-Species Hyperspectral Fish Freshness Assessment}


\maketitle

\begin{abstract}
Hyperspectral models of fish freshness rarely transfer between species, and labelling a new species is slow and destructive. Freshness is also ordinal,
and standard domain alignment can erase that order. We present InvarOrd, which
adapts to a new species from $30$ labelled cubes ($5$ per rank) in
three stages: self-supervised spectral pretraining on unlabelled source cubes,
cross-species alignment by a shrinkage covariance discrepancy computed only
within matched ranks, and ordinal distillation into a compact student. The
alignment term is provably invariant to the rank means, so it gives no incentive to collapse the
ordinal axis. On paired salmon and cod data ($800$ cubes
each, $16$ days), a source-trained teacher reaches $1.6$ to $1.9$ days
mean absolute error (MAE) on the other species, with quadratic weighted kappa
(QWK) below zero. The $423$K-parameter student, $2.54\times$ smaller than its
teacher, attains the lowest MAE ($1.476$ days) and highest $\pm1$-day accuracy
($56.1\%$) among $14$ configurations, matching the best of six
ImageNet-pretrained backbones (ConvNeXt-T, $1.503$ days) at $66\times$ fewer
parameters, $1.44$\,ms per cube on a GPU and $5$\,s re-adaptation. InvarOrd beats
target-only training by $0.29$ days ($p=0.031$). Among alignment
objectives, rank-invariant alignment retains the most ordinal structure
(embedding rank correlation $0.183$, QWK $0.112$), whereas global
discrepancy lowers the rank correlation to $0.049$ and QWK to zero at only $0.014$ days higher MAE.
\end{abstract}


\section{Introduction}
\label{sec:intro}

\IEEEPARstart{F}ish spoils continuously during cold storage, and knowing how long a fillet has been stored drives shelf-life labelling, recall decisions and the price
paid along the supply chain. Hyperspectral imaging (HSI) can estimate this
non-destructively, because protein breakdown, lipid oxidation and water loss
shift the reflectance spectrum in step with storage time
\cite{cheng2014tifs,wu2013talanta}. Deep spectral and spectral--spatial
networks have made such estimates practical on single species
\cite{salmon2025cfi,alam2026sgnet}, and compact models now support non-destructive agrifood inspection more widely \cite{karmakar2023wheat,oyinlola2026rf,wang2025plum,padeiro2026spq}. Yet three properties of the task still
defeat generic HSI classifiers.

First, freshness labels are \emph{ordinal}: confusing day two with day ten is
far worse than confusing day two with day three, and plain cross-entropy
ignores this \cite{cao2020coral,shi2023corn}. Second, labels are
\emph{scarce}, since each fillet must be tracked over days and graded
destructively, which places a new species firmly in the few-shot regime
\cite{alam2026fewshot,hsisurvey2025}. Third, models do not \emph{transfer
across species}. Salmon and cod differ in fat, pigmentation and spectral
shape, and a teacher trained on one loses the freshness ordering on the
other. The usual remedy, distribution
alignment \cite{sun2016deepcoral,ganin2015dann}, matches source and target
globally, and in doing so it can pull together samples of different
freshness and erase the very order the task depends on. Meanwhile the
strongest small-data accuracy often comes from ImageNet-scale backbones
\cite{liu2022convnext,dosovitskiy2021vit}, too costly for an inspection line.

We propose InvarOrd, built on one principle: cross-species alignment should
respect the ordinal structure of freshness rather than erase it. It extends
our lightweight spectral backbone \cite{alam2026sgnet} and few-shot ordinal
formulation \cite{alam2026fewshot} from one species to two, and spends the
small label budget only where labels are indispensable. The encoder is first
pretrained on unlabelled source cubes, so the $30$ labelled target cubes are
used for transfer rather than for learning spectral features from scratch. A
covariance discrepancy \cite{sun2016deepcoral} is then computed only between
source and target samples of the same freshness rank. Because each rank is
centred separately, the alignment can match how the two species vary within a
rank without being able to pull ranks towards one another. Finally, the
aligned teacher is distilled into a compact ordinal student
\cite{cao2020coral} through its full cumulative rank distribution rather than
hard labels. Our aim is not to outscore ImageNet-scale backbones, whose best
error is statistically tied with ours, but to match them at a fraction of the
cost while keeping the ordinal structure that global alignment destroys.

The contributions of this article are as follows.
\begin{itemize}
\item A rank-invariant alignment that matches covariances only within the same freshness rank, with a shrinkage estimator for the $k$-shot regime. Beyond per-class pooling, we show that the statistic matters for ordered classes, and prove that, unlike global or mean-matching objectives, this term is invariant to the rank means and so cannot reduce its own value by collapsing the ordinal axis.
\item A three-stage pipeline of self-supervised spectral pretraining on
unlabelled source cubes, rank-invariant alignment and ordinal knowledge
distillation, which yields a
$423$K-parameter student that adapts to a new species from $30$ labelled
cubes in about five seconds.
\item An evaluation against six ImageNet-pretrained backbones under linear
probing and full fine-tuning, with paired significance tests. The compact student matches the best backbone at $66\times$ fewer parameters, and among
alignment objectives rank-invariant alignment retains the most ordinal
structure, while global MMD drives quadratic weighted kappa to zero at
near-identical error.
\end{itemize}

Section~\ref{sec:related} reviews related work, Section~\ref{sec:method} presents InvarOrd and the rank-invariance analysis, Section~\ref{sec:results} reports the experiments, Section~\ref{sec:discussion} discusses the findings and their limits, and Section~\ref{sec:conclusion} concludes.
\section{Related Work}
\label{sec:related}

\subsection{Hyperspectral Imaging for Fish Freshness}
HSI has been used to assess fish quality for over a decade, first with
chemometric regression of freshness indicators such as total volatile basic
nitrogen and microbial load \cite{cheng2014tifs,wu2013talanta}, and more
recently with fused spectral indices and deep models for salmon
\cite{salmon2025cfi}. Imaging spectroscopy has mapped fat and water in fillets on a conveyor \cite{elmasry2008online}, hyperspectral sensing has been reviewed for wheat quality and plant disease \cite{karmakar2023wheat,rayhana2023plant}, band reduction has made hyperspectral rice-seed classification cheaper \cite{fabiyi2024rice}, and machine learning has improved other non-destructive sensors \cite{oyinlola2026rf}. Our earlier work introduced a lightweight
spectral-grouped classifier \cite{alam2026sgnet}, a few-shot ordinal
formulation for day-wise estimation \cite{alam2026fewshot}, and the public
HFQA dataset \cite{alam2026hfqa}. This literature is almost entirely
single-species and fully supervised, and none of it transfers a model to a
new species under a small label budget; InvarOrd keeps their compact, non-destructive spectral input and adds this transfer.

\subsection{Deep Spectral--Spatial Architectures}
Three-dimensional and hybrid convolutional networks established the
spectral--spatial paradigm for HSI \cite{zhong2017ssrn,roy2019hybridsn}, and
transformers later treated the spectrum as a sequence
\cite{hong2022spectralformer,sun2022ssftt}. Recent designs add deformable
convolution and cluster attention \cite{fang2025dcht} or lighten the attention
for deployment \cite{ma2023lsga}. A parallel route adapts general-purpose
ImageNet backbones, such as ResNet \cite{he2016resnet}, EfficientNet
\cite{tan2019efficientnet}, ConvNeXt \cite{liu2022convnext}, ViT
\cite{dosovitskiy2021vit} and Swin \cite{liu2021swin}, through a spectral
adapter. These backbones carry tens of millions of parameters, and their best
adaptation regime changes from model to model at $30$ labels. We use them as
external baselines and pursue the opposite design point, a compact encoder
trained end to end at a scale suited to the label budget.

\subsection{Ordinal Regression}
Ordinal regression decomposes a $K$-level label into $K-1$ cumulative binary
problems \cite{frank2001ordinal}, an idea carried into deep networks for age
estimation \cite{niu2016ordinal}. CORAL ties these binary tasks through a
shared weight vector to guarantee rank-consistent outputs \cite{cao2020coral},
and CORN relaxes the shared weight with conditional training
\cite{shi2023corn}. Later work models ordinal label distributions
\cite{wen2023oldl} and uses language priors through learnable rank prompts
\cite{li2022ordinalclip}. All assume a single domain: they define how a
classifier should respect order, not how order should survive a domain
change.

\subsection{Few-Shot and Cross-Domain HSI Classification}
Metric and optimization-based meta-learning \cite{snell2017proto,finn2017maml}
underpin much of few-shot HSI, including calibrated multi-view prototypes
\cite{yu2022multiview} and transductive embeddings \cite{xi2025tefsl}, and a
recent survey reviews the limited-sample problem in depth \cite{hsisurvey2025}. Cross-domain variants add meta-learning under dual
adjustment \cite{hu2023dualadjust}, domain-adversarial training
\cite{ye2023adafsl}, spectral coordinate transformers \cite{li2024scformer}
and transformer-based target detection \cite{feng2024tcdfsl}. All of these
are built for nominal land-cover or target classes. Their alignment terms
treat every class as equidistant, which is harmless but
damaging when the classes are ordered.

\subsection{Distribution Alignment and Ordinal Structure}
Unsupervised domain adaptation reduces a discrepancy between source and
target features \cite{wang2018survey}, through kernel mean embeddings
\cite{gretton2012mmd,long2015dan}, adversarial domain classifiers
\cite{ganin2015dann} or second-order statistics \cite{sun2016deepcoral}.
Covariance matching is attractive at small sample sizes, but the sample
covariance of a few anchors is singular, which shrinkage estimators correct
\cite{ledoit2004shrinkage}. Most adaptation methods align marginal
distributions, so a global objective is blind to which samples share a label.
Class-conditional alignment addresses this for nominal classes, for example by
contrasting intra-class and inter-class discrepancies across domains
\cite{kang2019can}, so pooling by class is not new in itself. What has not been examined is which statistic to match once classes are ordered: any objective that matches first moments, globally or per class, can lower its loss by moving class means, which then carry the freshness axis. For ordered classes the requirement is stronger, since
alignment should neither merge neighbouring ranks nor reorder distant ones,
and treating ranks as unrelated categories does not encode this.
Structure-preserving ordinal adaptation \cite{tian2024spoda} is the closest
work to ours: it keeps ordinal structure during unsupervised adaptation. We differ in working with a handful of labelled
target anchors, in restricting alignment to matched ranks with a shrinkage
covariance, and in proving an invariance property of the alignment term that explains why
it does not itself reward collapsing the ordinal axis.

\subsection{Self-Supervised Pretraining and Knowledge Distillation}
Masked reconstruction \cite{he2022mae}, used in agrifood for yield forecasting \cite{beddows2026visiontrees}, and contrastive learning
\cite{chen2020simclr} learn without labels, which suits HSI, where
unlabelled cubes are cheap and graded ones are not; Stage~1 masks along the spectral axis.
Knowledge distillation transfers a teacher's soft outputs to a smaller
student \cite{hinton2015kd}, through intermediate hints
\cite{romero2015fitnets} or between equal-size networks
\cite{furlanello2018born,gou2021kdsurvey}. Recent work refines what is transferred, using
orthogonal projections \cite{miles2024vkd} or Brownian distance covariance
\cite{lu2025bdckd}. Distillation for ordinal outputs has received little
attention. Because each CORAL output is a Bernoulli probability, the teacher's
full cumulative rank distribution can be passed to the student, which is how
our student keeps the teacher's accuracy at $2.54\times$ fewer parameters.

\subsection{\texorpdfstring{Lightweight Models for Agrifood Inspection}{Lightweight Models for Agrifood Inspection}}
\label{subsec:agrifood}
Distillation has produced lightweight defect classifiers for green plums \cite{wang2025plum} and TinyML weed classifiers for IoT nodes \cite{zhang2026tinyml}; hardware-aware design runs weed identification on-device \cite{zhang2026ondevice} and on low-power microcontrollers \cite{li2026wagenet}; improved detectors grade tomato maturity in greenhouses \cite{chen2026yolorcm}; and similarity-preserving quantisation makes crop disease detection portable \cite{padeiro2026spq}. These works share our aim of accurate inference on a tight budget but use RGB input, nominal classes and no population shift. InvarOrd adds hyperspectral input, a distilled ordinal output and adaptation to a new species from a few labelled cubes, a lightweight form of the model updating that continual learning addresses in agrifood \cite{dimitri2026continual}.

As Table~\ref{tab:positioning} shows, each line of work covers only part of
the problem: ordinal methods assume one domain, cross-domain HSI methods ignore
order, and ordinal adaptation has been studied without labelled target
anchors and outside spectral imaging. To our knowledge InvarOrd is the first
method to address all four properties together.

\begin{table}[htbp]
\centering
\caption{Positioning against representative prior work. Few-shot: labelled
target anchors; Cross-dom.: cross-domain transfer;}
\label{tab:positioning}
\footnotesize
\setlength{\tabcolsep}{2.5pt}
\begin{tabular}{@{}lcccc@{}}
\toprule
Method & Ordinal & Few-shot & Cross-dom. & HSI\\
\midrule
CORAL, CORN \cite{cao2020coral,shi2023corn} & \checkmark & -- & -- & --\\
Salmon freshness index \cite{salmon2025cfi} & -- & -- & -- & \checkmark\\
Few-shot ordinal HSI \cite{alam2026fewshot} & \checkmark & \checkmark & -- & \checkmark\\
Few-shot HSI \cite{yu2022multiview,xi2025tefsl} & -- & \checkmark & -- & \checkmark\\
Cross-domain FS-HSI \cite{hu2023dualadjust,ye2023adafsl,li2024scformer} & -- & \checkmark & \checkmark & \checkmark\\
Deep CORAL, DANN, CAN \cite{ganin2015dann,kang2019can} & -- & -- & \checkmark & --\\
SPODA \cite{tian2024spoda} & \checkmark & -- & \checkmark & --\\
\midrule
\textbf{InvarOrd (ours)} & \checkmark & \checkmark & \checkmark & \checkmark\\
\bottomrule
\end{tabular}
\end{table}

\section{Methodology}
\label{sec:method}

\begin{figure*}[t]
\centering
\includegraphics[width=\textwidth]{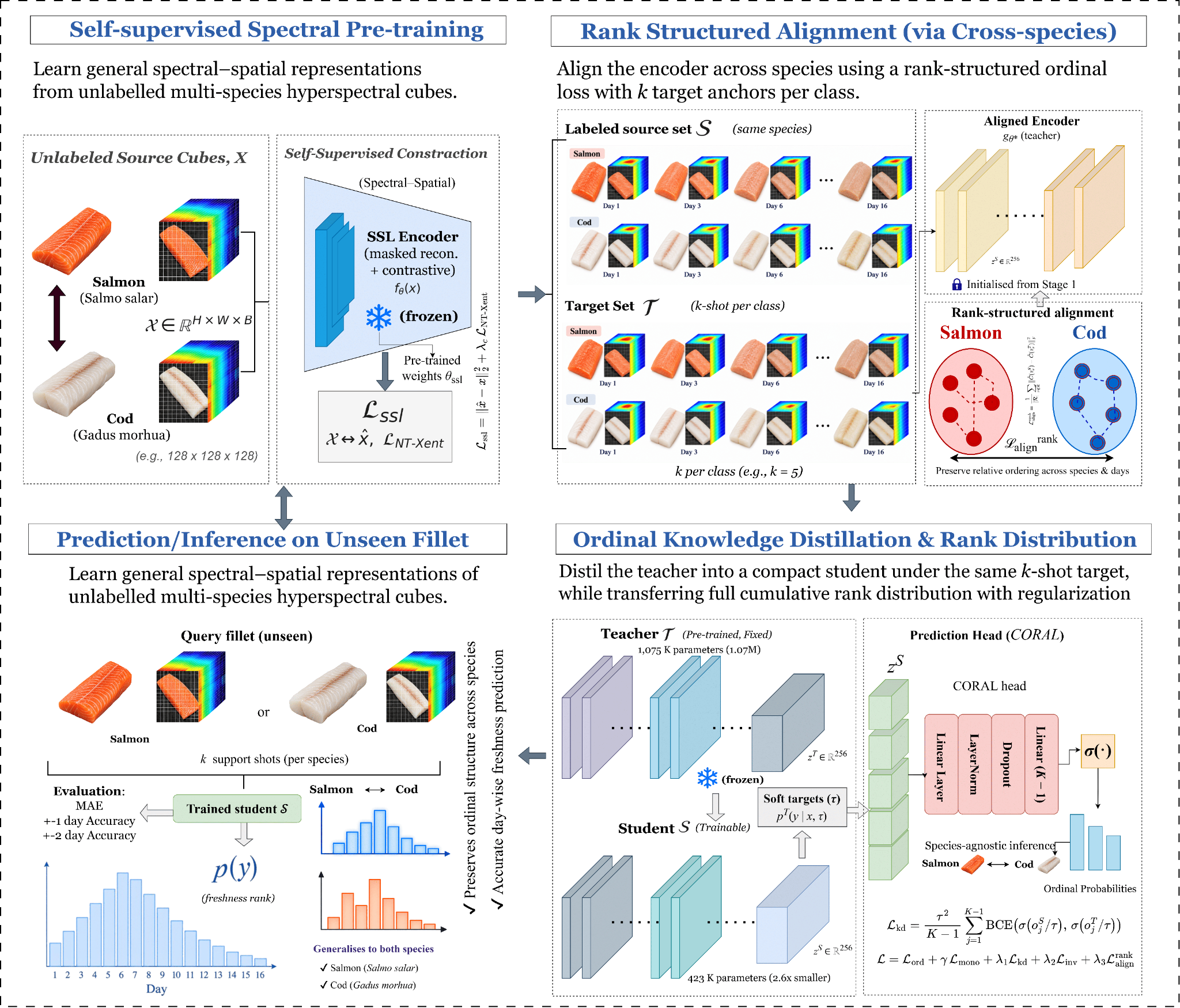}
\caption[Overview of InvarOrd]{Overview of InvarOrd, three training stages and deployment: (1) self-supervised spectral pre-training on unlabelled source-species hyperspectral cubes (salmon or cod, depending on the transfer direction); (2) rank-structured cross-species alignment using $k$ target anchors per class; (3) ordinal knowledge distillation of the teacher's cumulative rank distribution into a compact student; and (4) freshness prediction on unseen target-species fillets.}
\label{fig:pipeline}
\end{figure*}

InvarOrd learns a compact ordinal freshness predictor for a target species
from a small labelled budget by transferring the ordinal structure learnt on a
data-rich source species. Training proceeds in three stages: self-supervised
spectral pretraining (Stage~1), rank-structured cross-species alignment that
yields the teacher (Stage~2), and ordinal knowledge distillation into the
deployed student (Stage~3). The full procedure is summarised in
Fig.~\ref{fig:pipeline} and Algorithm~\ref{alg:invarord}.

\subsection{Problem Formulation}
\label{subsec:problem}

Let $\mathcal{S}$ and $\mathcal{T}$ denote the source and target species
(salmon and cod, transferred in both directions). Each sample is a
preprocessed hyperspectral cube $\mathbf{x}\in\mathbb{R}^{B\times H\times W}$
with $B=64$ bands, uniformly subsampled from $462$ raw bands spanning
$400$ to $1000$\,nm, and $H=W=32$. Storage days are discretised into $K$
ordered freshness ranks through fixed bin edges
$e_0<e_1<\dots<e_K$,
\begin{equation}
y \;=\; r \quad\Longleftrightarrow\quad e_r \le \mathrm{day} < e_{r+1},
\qquad r\in\{0,\dots,K\!-\!1\}.
\label{eq:binning}
\end{equation}
The source provides a labelled set
$\mathcal{D}_{\mathcal S}=\{(\mathbf{x}_i^{\mathcal S},y_i^{\mathcal S})\}_{i=1}^{N_{\mathcal S}}$
and an unlabelled pool $\mathcal{U}_{\mathcal S}$, whereas the target supplies
only a $k$-shot anchor set
\begin{equation}
\mathcal{D}_{\mathcal T}^{k}
=\bigcup_{r=0}^{K-1}\big\{(\mathbf{x}_j^{\mathcal T},r)\big\}_{j=1}^{k},
\qquad |\mathcal{D}_{\mathcal T}^{k}| = kK,
\label{eq:anchors}
\end{equation}
which gives $30$ labelled cubes at the default $k=5$, $K=6$. No further target
label is used at any stage. The aim is a student
$f_{\mathrm s}:\mathbb{R}^{B\times H\times W}\!\to\!\{0,\dots,K\!-\!1\}$ whose
parameter and compute footprint suit on-instrument deployment, while its
ordinal error on $\mathcal{T}$ remains competitive with far larger pretrained
backbones.


\subsection{Spectral--Spatial Backbone and Ordinal Head}
\label{subsec:backbone}

Teacher and student share one encoder family, written as the composition
\begin{equation}
g_\theta(\mathbf{x})
= \mathbf{P}\,\mathrm{GAP}\big(\mathcal{B}_L\circ\cdots\circ\mathcal{B}_1
\circ\mathcal{E}(\mathbf{x})\big)\in\mathbb{R}^{d},
\label{eq:encoder}
\end{equation}
where $\mathcal{E}$ is a $1\times1$ spectral stem that mixes the $B$ bands into
$w$ channels, each block $\mathcal{B}_l$ applies a standard $3\times3$
convolution followed by a depthwise $3\times3$ convolution, each with
GroupNorm and GELU, and then $2\times2$ max pooling, $\mathrm{GAP}$ is global
average pooling and $\mathbf{P}$ is a linear projection to $d$ dimensions. We
use $L=4$ blocks throughout, so that width $w$ and embedding size $d$ are the
only capacity controls (Table~\ref{tab:arch}).

\vspace{-5pt}
\begin{table}[htbp]
\centering
\caption{Teacher and student configurations. Both share the encoder of
Eq.~\eqref{eq:encoder} with $L=4$ blocks and a CORAL head.}
\label{tab:arch}
\footnotesize
\setlength{\tabcolsep}{4pt}
\begin{tabular}{@{}lcccc@{}}
\toprule
Model & $(w,d)$ & Params & MACs & Role \\
\midrule
Teacher                & $(64,256)$ & $1.07$M  & $90.3$M & Stages 1 and 2 \\
Student (compressed)   & $(40,160)$ & $423$K   & $36.5$M & Stage 3, deployed \\
Student (same capacity)& $(64,256)$ & $1.07$M  & $90.3$M & Stage 3, deployed \\
\midrule
\multicolumn{5}{l}{Compression: $2.54\times$ parameters, $2.47\times$ MACs.} \\
\bottomrule
\end{tabular}
\end{table}

Because freshness is ordinal, prediction uses a CORAL head~\cite{cao2020coral}
with a single shared weight vector $\mathbf{u}\in\mathbb{R}^{d}$ and $K-1$
biases,
\begin{equation}
o_j(\mathbf{x})=\mathbf{u}^{\!\top}g_\theta(\mathbf{x})+b_j,
\qquad j=1,\dots,K-1,
\label{eq:coralhead}
\end{equation}
so that $p_j=\sigma(o_j)$ estimates $P(y>j)$ and the rank follows by counting
the thresholds that are passed,
\begin{equation}
\hat{y}=\sum_{j=1}^{K-1}\mathbb{I}\big[p_j>\tfrac12\big].
\label{eq:decode}
\end{equation}
With extended binary targets $t_j=\mathbb{I}[y>j]$, the ordinal loss is
\begin{equation}
\mathcal{L}_{\mathrm{ord}}(\mathbf{x},y)
=\frac{1}{K-1}\sum_{j=1}^{K-1}\mathrm{BCE}\big(o_j(\mathbf{x}),t_j\big).
\label{eq:coralloss}
\end{equation}
Sharing $\mathbf{u}$ already makes the logits ordered whenever the biases are
ordered, but the biases are free parameters and may cross during few-shot
updates. We therefore add a hinge penalty on the cumulative probabilities,
\begin{equation}
\mathcal{L}_{\mathrm{mono}}
=\frac{1}{K-2}\sum_{j=1}^{K-2}\big[p_{j+1}-p_j\big]_{+},
\label{eq:mono}
\end{equation}
where $[a]_+=\max(0,a)$. It vanishes exactly when $p_j$ is non-increasing in
$j$, which is required for a valid ordinal model.

\subsection{Stage 1: Self-Supervised Spectral Pretraining}
\label{subsec:ssl}

The encoder is first pretrained on $\mathcal{U}_{\mathcal S}$, which costs no
labels. A random band mask $\mathbf{M}\in\{0,1\}^{B}$ with masking ratio
$m=0.5$ is broadcast across space to give
$\tilde{\mathbf{x}}=\mathbf{M}\odot\mathbf{x}$, and a light decoder $h_\phi$
reconstructs the full cube, $\hat{\mathbf{x}}=h_\phi(g_\theta(\tilde{\mathbf x}))$.
A contrastive term is added on two augmented views of each cube. For a batch
of $N$ cubes giving $2N$ normalised embeddings, with $(a,a^{+})$ a positive
pair and temperature $\kappa$,
\begin{equation}
\mathcal{L}_{\mathrm{NT\text{-}Xent}}
=-\frac{1}{2N}\sum_{a=1}^{2N}
\log\frac{\exp\big(\mathbf{z}_a^{\!\top}\mathbf{z}_{a^{+}}/\kappa\big)}
{\sum_{b\neq a}\exp\big(\mathbf{z}_a^{\!\top}\mathbf{z}_{b}/\kappa\big)}.
\label{eq:ntxent}
\end{equation}
The Stage-1 objective is
\begin{equation}
\mathcal{L}_{\mathrm{ssl}}
=\big\|\hat{\mathbf{x}}-\mathbf{x}\big\|_2^2
+\lambda_{\mathrm c}\,\mathcal{L}_{\mathrm{NT\text{-}Xent}},
\label{eq:ssl}
\end{equation}
with $\lambda_{\mathrm c}=0.3$. Reconstruction forces the encoder to model
inter-band dependencies, which carry the chemical signature of spoilage, while
the contrastive term makes the embedding stable under nuisance augmentation.
The result is the spectral prior $\theta_{\mathrm{ssl}}$.

\subsection{Stage 2: Rank-Structured Cross-Species Alignment}
\label{subsec:align}

Initialised from $\theta_{\mathrm{ssl}}$, the teacher is trained on
$\mathcal{D}_{\mathcal S}$ together with the anchors
$\mathcal{D}^{k}_{\mathcal T}$. Alongside the ordinal loss, a second-order
discrepancy aligns the two species in embedding space. Within a rank, only
$k=5$ target samples are available against $d=256$ dimensions, so the sample
covariance has rank at most $k-1$ and is badly conditioned. We therefore use a
shrinkage estimator on centred embeddings
$\bar{\mathbf{Z}}=(\mathbf{I}_n-\tfrac1n\mathbf{1}\mathbf{1}^{\!\top})\mathbf{Z}$,
\begin{equation}
\widehat{\mathbf C}(\mathbf Z)
=(1-\alpha)\frac{\bar{\mathbf Z}^{\!\top}\bar{\mathbf Z}}{n-1}
+\alpha\,\frac{\mathrm{tr}(\bar{\mathbf Z}^{\!\top}\bar{\mathbf Z})}{d\,(n-1)}\,
\mathbf I_d,
\label{eq:shrinkcov}
\end{equation}
with $\alpha=0.1$, and the CORAL discrepancy~\cite{sun2016deepcoral}
\begin{equation}
\mathcal{L}_{\mathrm{align}}(\mathbf Z_{\mathcal S},\mathbf Z_{\mathcal T})
=\frac{1}{4d^{2}}
\big\|\widehat{\mathbf C}(\mathbf Z_{\mathcal S})
-\widehat{\mathbf C}(\mathbf Z_{\mathcal T})\big\|_F^{2}.
\label{eq:coral}
\end{equation}
The defining choice of InvarOrd is to evaluate Eq.~\eqref{eq:coral} only
between samples of the same freshness rank and to average over the ranks
$\mathcal{R}$ with at least two samples in both domains of a batch,
\begin{equation}
\mathcal{L}_{\mathrm{align}}^{\mathrm{rank}}
=\frac{1}{|\mathcal R|}\sum_{r\in\mathcal R}
\mathcal{L}_{\mathrm{align}}\big(\mathbf Z^{(r)}_{\mathcal S},
\mathbf Z^{(r)}_{\mathcal T}\big),
\label{eq:rankalign}
\end{equation}
since Eq.~\eqref{eq:shrinkcov} needs $n\ge2$; at $k=1$ each rank has one anchor, so this term is inactive and the other terms adapt the model.
The Stage-2 objective is
\begin{equation}
\mathcal{L}_{\mathrm{II}}
=\mathcal{L}_{\mathrm{ord}}^{\mathcal S}
+\lambda_{\mathrm a}\,\mathcal{L}_{\mathrm{align}}^{\mathrm{rank}},
\label{eq:stage2}
\end{equation}
where $\mathcal{L}_{\mathrm{ord}}^{\mathcal S}$ is Eq.~\eqref{eq:coralloss}
averaged over source samples. Its minimiser gives the teacher encoder
$g_{\theta^\star}$ and head; the anchors enter Stage~2 only through the
alignment term.

The motivation for matching within ranks rather than globally can be made
precise. By the law of total covariance, the population covariance of a
domain decomposes as
\begin{equation}
\mathbf C_{\mathrm{tot}}
=\underbrace{\textstyle\sum_{r}\pi_r\,\mathbf C_r}_{\text{within-rank}}
+\underbrace{\textstyle\sum_{r}\pi_r(\bm\mu_r-\bm\mu)(\bm\mu_r-\bm\mu)^{\!\top}}_{\boldsymbol{\Sigma}_{\mathrm B}:\ \text{between-rank}},
\label{eq:totalcov}
\end{equation}
where $\pi_r$, $\bm\mu_r$ and $\mathbf C_r$ are the prior, mean and covariance
of rank $r$. The ordinal information lies largely in $\boldsymbol\Sigma_{\mathrm B}$,
the spread of rank means along the freshness axis.

\begin{proposition}
\label{prop:invariance}
$\mathcal{L}_{\mathrm{align}}^{\mathrm{rank}}$ is invariant to any
translation $\mathbf Z^{(r)}\mapsto\mathbf Z^{(r)}+\mathbf 1\mathbf v_r^{\!\top}$
applied independently to each rank in either domain. Hence its gradient has
no component along the rank means, and it cannot reduce its own value by
contracting $\boldsymbol\Sigma_{\mathrm B}$.
\end{proposition}

\noindent\textit{Proof.} Centring removes any common row offset, since
$(\mathbf I_n-\tfrac1n\mathbf 1\mathbf 1^{\!\top})\mathbf 1=\mathbf 0$. Thus
$\bar{\mathbf Z}^{(r)}$, and with it both terms of
Eq.~\eqref{eq:shrinkcov}, is unchanged, and so is every
summand of Eq.~\eqref{eq:rankalign}. \hfill$\square$

\smallskip
We call this property rank invariance, hence the name InvarOrd. A global
discrepancy has no such property. It matches
$\mathbf C_{\mathrm{tot}}$, which contains $\boldsymbol\Sigma_{\mathrm B}$,
so when the two species differ in how widely their ranks spread, the loss can
be lowered by shrinking the rank separation in one of them. The matched-rank term offers no such incentive. The proposition is a
statement about the alignment term alone, not a guarantee for the whole training process; the ordinal terms in
Eq.~\eqref{eq:stage2} remain responsible for actively separating the ranks, and other terms, such as $\mathcal{L}_{\mathrm{inv}}$ in Stage~3, are not covered by it.
Table~\ref{tab:alignvariants} contrasts the alignment choices examined in
this work.

Proposition~\ref{prop:invariance} separates two choices that are easily conflated. Pooling by rank, shared with class-conditional alignment, does not by itself give invariance: the rank-pooled MMD of Table~\ref{tab:alignvariants} also matches rank means through its kernel. Invariance needs per-rank pooling \emph{and} a statistic of centred embeddings; Section~\ref{subsec:ablation} tests the two separately.

\begin{table}[t]
\centering
\caption{Alignment objectives compared in this work.}
\label{tab:alignvariants}
\footnotesize
\setlength{\tabcolsep}{2.5pt}
\begin{tabular}{@{}lccc@{}}
\toprule
Alignment & Statistic & Pooling & Invariant to rank means \\
\midrule
Global MMD (RBF)        & all moments   & global   & no  \\
Adversarial (DANN-type) & domain label  & global   & no  \\
Global CORAL            & covariance    & global   & no  \\
Class-cond.\ \cite{kang2019can} & all moments & per class & no \\
Rank-pooled MMD         & all moments   & per rank & no  \\
\textbf{Rank CORAL (ours)} & covariance & per rank & \textbf{yes} \\
\bottomrule
\end{tabular}
\end{table}

\subsection{Stage 3: Ordinal Knowledge Distillation}
\label{subsec:kd}

The deployed predictor is a student $f_{\mathrm s}$ trained under the same
$kK$ anchor budget, with the frozen teacher $f_{\mathrm t}$ providing soft
ordinal targets. The student encoder is initialised from the Stage-2 teacher
wherever tensor shapes agree; tensors of any other shape cannot inherit the teacher and are initialised independently of it,
\begin{equation}
\theta_{\mathrm s}^{(\ell)}\leftarrow
\begin{cases}
\theta^{\star(\ell)}, & \text{if } \dim\theta_{\mathrm s}^{(\ell)}=\dim\theta^{\star(\ell)},\\[2pt]
\text{independent init.}, & \text{otherwise,}
\end{cases}
\label{eq:inherit}
\end{equation}
for each parameter tensor $\ell$. When every tensor is inherited, as in the
same-capacity student, the stem and the lower blocks are frozen so that the
aligned geometry is retained and only the upper blocks and head adapt. In the compressed student almost every tensor differs in shape, so the aligned geometry cannot reach it through the weights; it reaches it only through the Stage-3 losses below. We state this explicitly because the
two configurations inherit the alignment by different routes.

The distillation term matches the student's cut-point probabilities to the
teacher's at temperature $\tau$,
\begin{equation}
\mathcal{L}_{\mathrm{kd}}
=\frac{\tau^{2}}{K-1}\sum_{j=1}^{K-1}
\mathrm{BCE}\Big(o^{\mathrm s}_j/\tau,\;\sigma\big(o^{\mathrm t}_j/\tau\big)\Big).
\label{eq:kd}
\end{equation}
Each CORAL output is a Bernoulli probability, so Eq.~\eqref{eq:kd} transfers
the teacher's full cumulative rank distribution rather than a single hard
label, which is what makes distillation useful when only $30$ labelled cubes
exist. Since $o^{\mathrm s}_j=\mathbf u_{\mathrm s}^{\!\top}\mathbf z_{\mathrm s}+b^{\mathrm s}_j$, the gradient of Eq.~\eqref{eq:kd} with respect to the student embedding is
\begin{equation}
\frac{\partial\mathcal{L}_{\mathrm{kd}}}{\partial\mathbf z_{\mathrm s}}
=\frac{\tau}{K-1}\sum_{j=1}^{K-1}\Big(\sigma\big(o^{\mathrm s}_j/\tau\big)
-\sigma\big(o^{\mathrm t}_j/\tau\big)\Big)\,\mathbf u_{\mathrm s},
\label{eq:kdgrad}
\end{equation}
a multiple of $\mathbf u_{\mathrm s}$, as are those of $\mathcal{L}_{\mathrm{ord}}$ and $\mathcal{L}_{\mathrm{mono}}$. These terms depend on $\mathbf z_{\mathrm s}$ only through $\mathbf u_{\mathrm s}^{\!\top}\mathbf z_{\mathrm s}$ and do not constrain the orthogonal components, which hold the within-rank geometry aligned by Eq.~\eqref{eq:rankalign}. Eq.~\eqref{eq:kd} is also evaluated only on source cubes and the $kK$ anchors, so the teacher is never queried on other target cubes.

A domain-invariance regulariser uses an RBF-kernel MMD between source
and target-anchor embeddings,
\begin{equation}
\mathcal{L}_{\mathrm{inv}}
=\big\|\mathbb{E}_{\mathcal S}[\phi(\mathbf z)]
-\mathbb{E}_{\mathcal T}[\phi(\mathbf z)]\big\|_{\mathcal H}^{2},
\quad
\kappa(\mathbf z,\mathbf z')=e^{-\|\mathbf z-\mathbf z'\|^2/2\beta^2},
\label{eq:mmd}
\end{equation}
estimated with the unbiased U-statistic. Both $\mathcal{L}_{\mathrm{inv}}$ and
the matched-rank term are introduced gradually with a cosine warm-up,
\begin{equation}
\omega(t)=\tfrac12\Big(1-\cos\big(\pi\min(t/T_{\mathrm w},1)\big)\Big),
\label{eq:warmup}
\end{equation}
so that early updates are dominated by the supervised and distillation
signals. The Stage-3 objective at step $t$ is
\begin{equation}
\begin{aligned}
\mathcal{L}_{\mathrm{III}}
={}&\mathcal{L}_{\mathrm{ord}}
+\gamma\,\mathcal{L}_{\mathrm{mono}}
+\lambda_1\,\mathcal{L}_{\mathrm{kd}}\\
&+\omega(t)\big(\lambda_2\,\mathcal{L}_{\mathrm{inv}}
+\lambda_{\mathrm a}\,\mathcal{L}_{\mathrm{align}}^{\mathrm{rank}}\big),
\end{aligned}
\label{eq:total}
\end{equation}
with $\gamma=0.1$, $\lambda_1=0.5$, $\lambda_2=0.3$ and $\tau=1$. The $1/(4d^2)$ factor of Eq.~\eqref{eq:coral} removes the dependence on $d$, and the $\tau^2$ factor of Eq.~\eqref{eq:kd} keeps the distillation gradient roughly independent of $\tau$ \cite{hinton2015kd}, so the weights keep their meaning across settings.
Table~\ref{tab:stages} summarises what each stage sees, trains and
optimises.

\begin{table}[t]
\centering
\caption{Summary of the three training stages.}
\label{tab:stages}
\small
\begin{tabular}{@{}lccc@{}}
\toprule
& Stage 1 & Stage 2 & Stage 3 \\
\midrule
Data            & $\mathcal U_{\mathcal S}$ & $\mathcal D_{\mathcal S}$, $\mathcal D^k_{\mathcal T}$ & $\mathcal D_{\mathcal S}$, $\mathcal D^k_{\mathcal T}$ \\
Target labels   & 0 & $kK$ & $kK$ (same) \\
Trained model   & encoder, decoder & teacher & student \\
Objective       & Eq.~\eqref{eq:ssl} & Eq.~\eqref{eq:stage2} & Eq.~\eqref{eq:total} \\
Schedule        & 40 epochs & 30 epochs & 200 steps \\
Output          & $\theta_{\mathrm{ssl}}$ & $\theta^\star$ & $f_{\mathrm s}$ \\
\bottomrule
\end{tabular}
\end{table}

\subsection{Training Procedure and Complexity}
\label{subsec:impl}

Stage 2 uses a learning rate of $5\times10^{-4}$. Stage 3 uses AdamW at
$10^{-3}$ with weight decay $10^{-4}$, gradient clipping at a norm of
$5.0$ and a cosine learning-rate schedule. Every experiment is run in both
transfer directions, salmon$\rightarrow$cod and cod$\rightarrow$salmon, over
several random seeds that fix the anchor draw and the initialisation, and
results are averaged over seeds and directions.

The matched-rank term adds little cost. For a batch with $n_r$ samples of rank
$r$, forming the two covariances and their Frobenius distance costs
$\mathcal O(n_r d^{2})$, so one evaluation of
Eq.~\eqref{eq:rankalign} costs $\mathcal O(n d^{2})$ with $n=\sum_r n_r$,
which matches a single global CORAL evaluation. No matrix inversion or
eigendecomposition is required, since Eq.~\eqref{eq:shrinkcov} enters only
through a Frobenius norm. The alignment, invariance and distillation terms
are used only during training; at inference the student evaluates
Eqs.~\eqref{eq:encoder} to~\eqref{eq:decode} alone, so its cost is the one
listed in Table~\ref{tab:arch}.

\begin{algorithm}[htbp]
\caption{InvarOrd training}
\label{alg:invarord}
\small
\begin{algorithmic}[1]
\REQUIRE $\mathcal U_{\mathcal S}$, $\mathcal D_{\mathcal S}$, $\mathcal D^k_{\mathcal T}$; weights $\lambda_{\mathrm c},\gamma,\lambda_{\mathrm a},\lambda_1,\lambda_2$; $\alpha$, $\tau$, $T_{\mathrm w}$
\ENSURE Compact ordinal student $f_{\mathrm s}$
\item[] \textit{Stage 1: self-supervised pretraining}
\FOR{each mini-batch $\{\mathbf x\}\subset\mathcal U_{\mathcal S}$}
  \STATE Sample band mask $\mathbf M$; set $\tilde{\mathbf x}\leftarrow\mathbf M\odot\mathbf x$
  \STATE Form two augmented views; compute $\mathcal L_{\mathrm{ssl}}$ by Eq.~\eqref{eq:ssl}
  \STATE Update $\theta,\phi$
\ENDFOR
\STATE $\theta_{\mathrm{ssl}}\leftarrow\theta$
\item[] \textit{Stage 2: rank-structured alignment (teacher)}
\STATE Initialise teacher encoder with $\theta_{\mathrm{ssl}}$
\FOR{each batch of $\mathcal D_{\mathcal S}$ paired with $\mathcal D^k_{\mathcal T}$}
  \STATE $\mathcal R\leftarrow$ ranks with $\ge2$ samples in both domains
  \FOR{$r\in\mathcal R$}
    \STATE Compute $\widehat{\mathbf C}(\mathbf Z^{(r)}_{\mathcal S})$, $\widehat{\mathbf C}(\mathbf Z^{(r)}_{\mathcal T})$ by Eq.~\eqref{eq:shrinkcov}
  \ENDFOR
  \STATE Compute $\mathcal L_{\mathrm{II}}$ by Eqs.~\eqref{eq:rankalign} and~\eqref{eq:stage2}; update teacher
\ENDFOR
\STATE $\theta^\star\leftarrow$ teacher encoder; freeze teacher $f_{\mathrm t}$
\item[] \textit{Stage 3: ordinal distillation (student)}
\STATE Initialise student by Eq.~\eqref{eq:inherit}; freeze lower blocks if fully inherited
\FOR{$t=1,\dots,T$}
  \STATE Draw anchor batch from $\mathcal D^k_{\mathcal T}$ and source batch from $\mathcal D_{\mathcal S}$
  \STATE Teacher logits $o^{\mathrm t}_j$ (no gradient); student logits $o^{\mathrm s}_j$
  \STATE Compute $\omega(t)$ by Eq.~\eqref{eq:warmup} and $\mathcal L_{\mathrm{III}}$ by Eq.~\eqref{eq:total}
  \STATE Clip gradient norm; AdamW update of student
\ENDFOR
\RETURN $f_{\mathrm s}$
\end{algorithmic}
\end{algorithm}

\subsection{Ablation Variants}
\label{subsec:variants}

To isolate each component, every variant in the experiments is obtained by
switching terms of Eqs.~\eqref{eq:ssl}, \eqref{eq:stage2} and~\eqref{eq:total}
on or off, with all other settings fixed (Table~\ref{tab:ablationdef}). This
keeps the comparison within one code path, so that differences can be
attributed to the switched term rather than to implementation detail.

\begin{table}[t]
\centering
\caption{Definition of the ablation variants. \checkmark: term active;
--: removed; R: rank-pooled MMD; G: global MMD; adv.: adversarial.}
\label{tab:ablationdef}
\footnotesize
\begin{tabular}{@{}lccccc@{}}
\toprule
Variant & SSL & Stage 2 & Rank align & KD & $\mathcal L_{\mathrm{inv}}$ \\
\midrule
\textbf{InvarOrd (full)}   & \checkmark & \checkmark & \checkmark & \checkmark & \checkmark \\
w/o SSL (supervised init.) & --         & \checkmark & \checkmark & \checkmark & \checkmark \\
w/o Stage 2                & \checkmark & --         & --         & \checkmark & \checkmark \\
Rank-pooled MMD + KD       & \checkmark & \checkmark & R          & \checkmark & \checkmark \\
Global MMD align + KD      & \checkmark & \checkmark & G          & \checkmark & \checkmark \\
Adversarial align + KD     & \checkmark & \checkmark & adv.       & \checkmark & \checkmark \\
w/o $\mathcal L_{\mathrm{inv}}$ & \checkmark & \checkmark & \checkmark & \checkmark & -- \\
Target-only scratch        & --         & --         & --         & --         & --         \\
\bottomrule
\end{tabular}
\end{table}

\section{Experiments and Results}
\label{sec:results}

We ask five questions: how large the cross-species gap is (Q1), how a
compact student adapted from $30$ target cubes compares with far larger
pretrained backbones (Q2), what compression costs (Q3), whether matched-rank
alignment preserves ordinal structure where global alignment does not (Q4),
and how performance scales with the target budget (Q5).

\subsection{Setup}
\label{subsec:protocol}

\begin{figure}[htbp]
\centering
\includegraphics[width=\columnwidth]{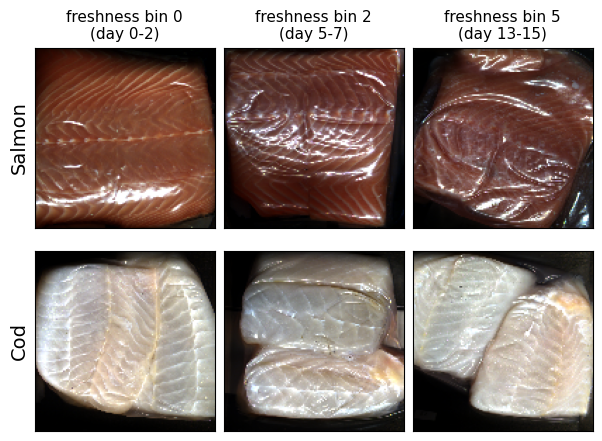}
\caption{Pseudo-RGB of salmon and cod fillets across the six
ranks.}
\label{fig:dataset}
\end{figure}

\begin{figure}[htbp]
\centering
\includegraphics[width=\columnwidth]{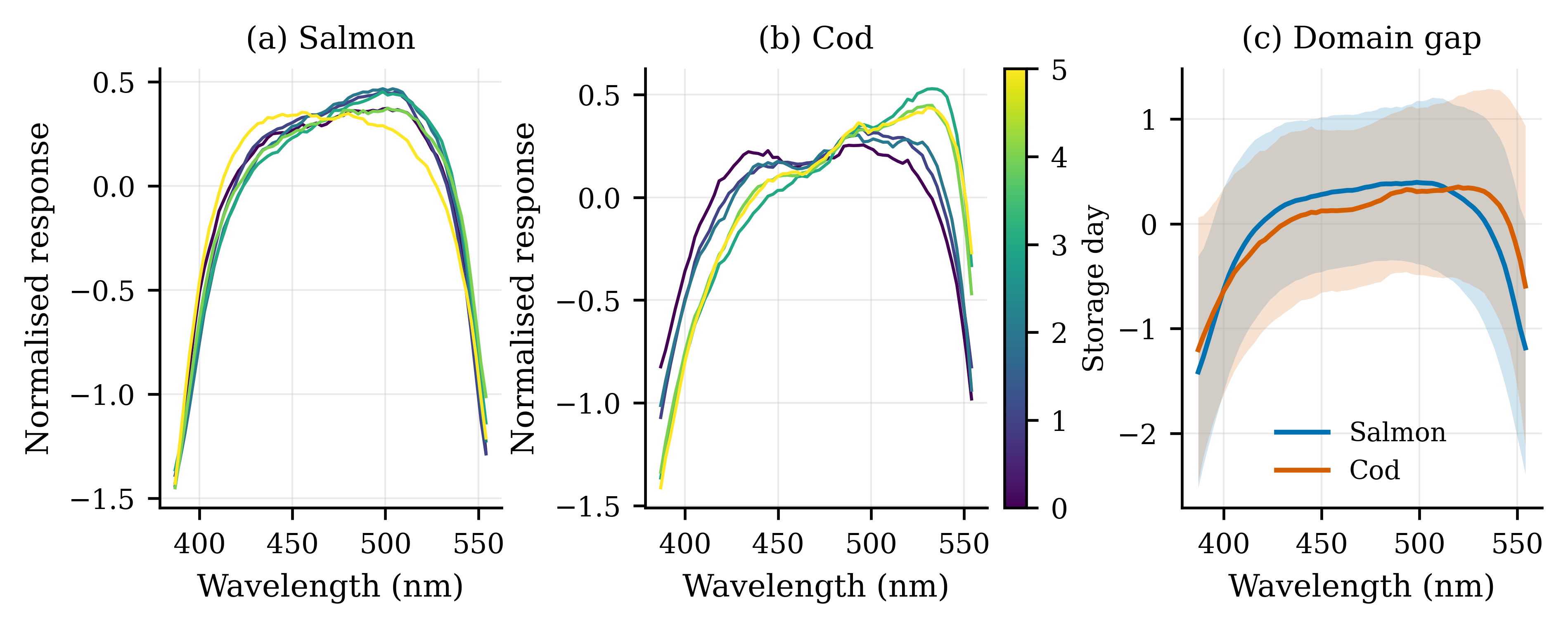}
\caption{Mean spectral response over storage day for (a) salmon and (b) cod, and
(c) the salmon--cod domain gap.}
\label{fig:spectra}
\end{figure}

Table~\ref{tab:protocol} summarises the data and protocol, and
Figs.~\ref{fig:dataset} and~\ref{fig:spectra} show the fillets and their
spectra. Spectra drift steadily with storage time within each species but
differ in level and shape between species, which makes transfer non-trivial.
Days are binned into $K=6$ ranks of about three days
(Eq.~\eqref{eq:binning}), since the per-day signal does not survive
preprocessing at $64\times32^2$, and all splits are pack-disjoint. Six
ImageNet-pretrained backbones with a spectral adapter are adapted on the same
$30$ target cubes by linear probing (LP) and full fine-tuning (FT); internal
baselines are listed in Table~\ref{tab:ablationdef}.

\emph{Freshness labels.} Fillets were kept in cold storage and imaged on each storage day \cite{alam2026hfqa}. Under constant chilled storage, total volatile basic nitrogen (TVB-N) and microbial counts rise with time \cite{gram1996spoilage}, and hyperspectral models of these indices are calibrated across a storage series \cite{cheng2014tifs,wu2013talanta}; EU rules place \emph{Salmo salar} and the \emph{Gadidae} in the same TVB-N category ($35\,$mg\,N/100\,g) \cite{eu2019627}. Because the CORAL head uses only label order, any index rising monotonically with storage time gives the same ranks up to the bin edges, and ranks are defined within each species, so equal spoilage rates are not assumed. Storage time remains a proxy that misses pack-to-pack variation and temperature abuse, which the three-day bins only partly absorb.


For $N$ test cubes with true storage day $d_i$ and predicted day $\hat d_i$,
the centre of the predicted rank $\hat y_i$,
\begin{equation}
\mathrm{MAE}=\frac{1}{N}\sum_{i=1}^{N}\big|\hat d_i-d_i\big|,
\label{eq:mae}
\end{equation}
\begin{equation}
\mathrm{Acc}_{\pm\delta}=\frac{1}{N}\sum_{i=1}^{N}
\mathbb{I}\big[\,|\hat d_i-d_i|\le\delta\,\big],
\label{eq:acc}
\end{equation}
and ordinal agreement is the quadratic weighted kappa
\begin{equation}
\mathrm{QWK}
=1-\frac{\sum_{a,b}(a-b)^2\,O_{ab}}{\sum_{a,b}(a-b)^2\,E_{ab}},
\label{eq:qwk}
\end{equation}
where $O$ and $E$ are the observed and chance rank confusion matrices, so
QWK $=0$ signals ordinal collapse. The embedding correlation $\rho$ is the
Spearman correlation between true rank and the CORAL score
$\mathbf{u}^{\!\top}\mathbf{z}$, and measures ordinal structure independently
of the thresholds. Paired tests treat each (direction, seed) combination as one pair and report the exact two-sided Wilcoxon $p$ \cite{demsar2006stats} and
$d_z=\overline{\Delta}/s_\Delta$, the mean paired MAE difference over its SD.
The smallest attainable exact $p$ is $2/2^{n}=0.031$ for $n=6$, however large the effect, and the two directions differ in difficulty (Table~\ref{tab:teacher}), so pairs are not identically distributed. We therefore read $d_z$ alongside $p$ and report directions separately where they disagree (Table~\ref{tab:students}).

\begin{table}[t]
\centering
\caption{Datasets and evaluation protocol.}
\label{tab:protocol}
\footnotesize
\begin{tabular}{@{}ll@{}}
\toprule
Item & Setting \\
\midrule
Species & salmon, cod; $800$ cubes and $50$ packs each \\
Storage period & $16$ days, binned into $K=6$ ranks \\
Cube (raw $\rightarrow$ used) & $462\times128^2\rightarrow64\times32^2$, $400$ to $1000$\,nm \\
Split & pack-disjoint, $30/10/10$ packs \\
Target budget & $k=5$ per rank ($30$ cubes); sweep $k\in\{1,3,5,10\}$ \\
Directions, seeds & $2$ directions $\times$ $3$ seeds, $n=6$ pairs \\
Accuracy metrics & MAE (days), $\mathrm{Acc}_{\pm1}$, $\mathrm{Acc}_{\pm2}$ \\
Ordinal metrics & QWK, embedding rank correlation $\rho$ \\
Significance & two-sided Wilcoxon signed-rank, Cohen's $d_z$ \\
\bottomrule
\end{tabular}
\end{table}

\subsection{Cross-Species Gap (Q1)}
\label{subsec:gap}

\begin{figure}[htbp]
\centering
\includegraphics[width=0.48\textwidth]{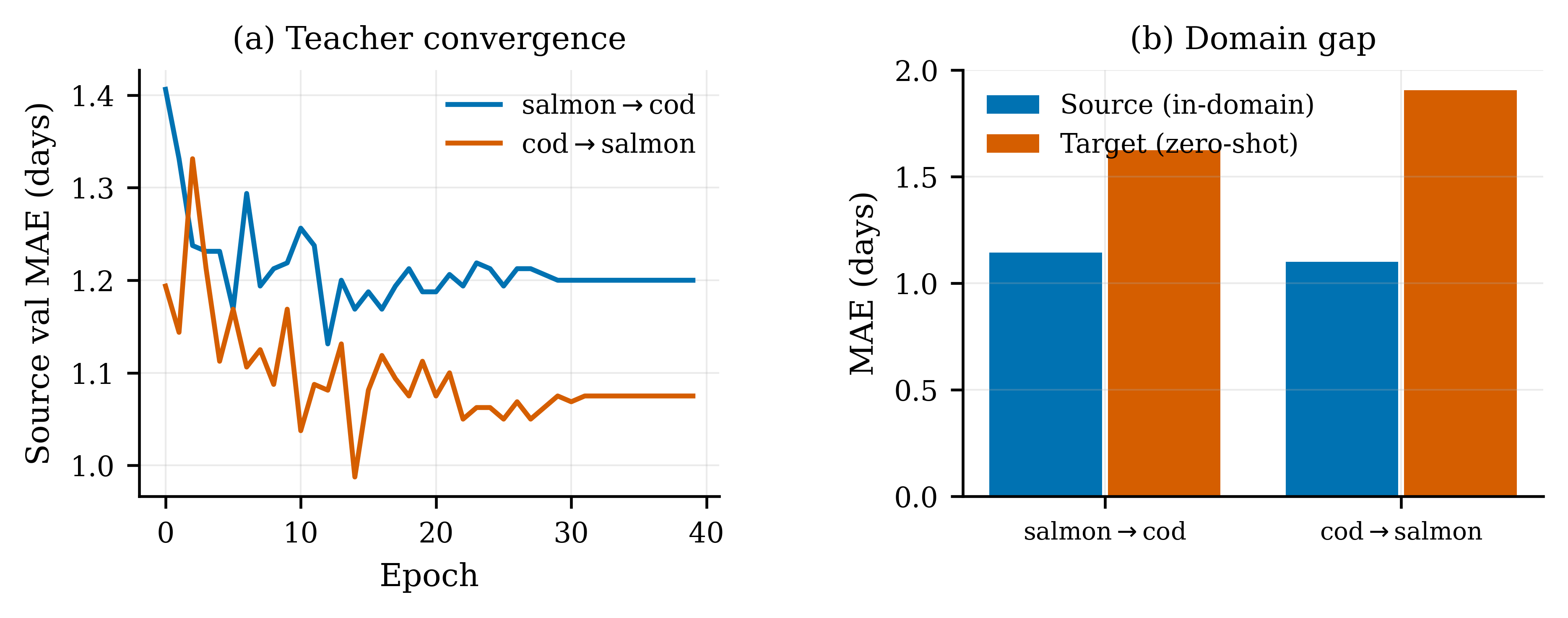}
\caption{Teacher source versus zero-shot target, both directions: (a) source
validation MAE during training; (b) source and zero-shot target MAE.}
\label{fig:teachergap}
\end{figure}

A teacher trained on one species and applied zero-shot to the other
degrades from about $1.1$ to between $1.6$ and $1.9$ days MAE, and its target
QWK falls below zero in both directions (Table~\ref{tab:teacher},
Fig.~\ref{fig:teachergap}). The source model therefore keeps some
sensitivity to spoilage but loses the freshness ordering on the unseen
species. This is the failure Stage~2 targets and the reference any few-shot
method must beat. The penalty is larger for cod$\rightarrow$salmon ($+0.81$
days) than for salmon$\rightarrow$cod ($+0.48$ days), and this asymmetry
recurs in every adapted result below.

\begin{table}[htbp]
\centering
\caption{Teacher on source and zero-shot on target.}
\label{tab:teacher}
\small
\begin{tabular}{@{}lccccc@{}}
\toprule
 & \multicolumn{2}{c}{Source} & \multicolumn{2}{c}{Target (zero-shot)} & \\
\cmidrule(lr){2-3}\cmidrule(lr){4-5}
Direction & MAE & QWK & MAE & QWK & $\Delta$MAE\\
\midrule
salmon$\rightarrow$cod & 1.144 & 0.500 & 1.625 & $-0.008$ & $+0.481$\\
cod$\rightarrow$salmon & 1.100 & 0.598 & 1.906 & $-0.044$ & $+0.806$\\
\bottomrule
\end{tabular}
\end{table}

\subsection{Comparison with Pretrained Backbones (Q2)}
\label{subsec:main}

Table~\ref{tab:main} reports all twelve backbone configurations beside both
students, and Table~\ref{tab:tradeoff} condenses each backbone to its
stronger regime. Both students give the lowest MAE ($1.476$ and $1.486$
days), and the compressed one gives the highest $\mathrm{Acc}_{\pm1}$
($56.1\%$). We read this with care. The top four methods lie within $0.057$
days, below the between-seed SD of InvarOrd ($0.134$ days), so the top of the
ranking is a statistical tie rather than a strict order. The defensible claim
is cost at equal accuracy: parity with the best backbone, ConvNeXt-T with
FT, at $65.7\times$ fewer parameters, and lower MAE than every ViT at $51$ to
$203\times$ fewer parameters.

\begin{table*}[t]
\centering
\caption{Comparison with ImageNet-pretrained backbones at $k=5$, mean over
directions and seeds. $\times$Params, $\times$MACs: relative to the compressed
student. Bold: best backbone value per metric; underline: stronger regime.
s$\rightarrow$c: salmon$\rightarrow$cod; c$\rightarrow$s: reverse.}
\label{tab:main}
\footnotesize
\setlength{\tabcolsep}{3pt}
\begin{tabular}{@{}llrrrrccccccc@{}}
\toprule
 & & & & & & \multicolumn{3}{c}{Linear probe} & \multicolumn{3}{c}{Full fine-tuning} & \\
\cmidrule(lr){7-9}\cmidrule(lr){10-12}
Method & Family & Params\,(M) & $\times$Params & MACs\,(M) & $\times$MACs
 & MAE\,$\downarrow$ & $\mathrm{Acc}_{\pm1}$\,$\uparrow$ & QWK\,$\uparrow$
 & MAE\,$\downarrow$ & $\mathrm{Acc}_{\pm1}$\,$\uparrow$ & QWK\,$\uparrow$ & FT$\,-\,$LP\\
\midrule
ConvNeXt-T      & CNN & 27.8 & 65.7  & 50  & 1.37 & 1.914 & 47.7 & 0.147 & \underline{\textbf{1.503}} & \underline{52.5} & \underline{0.094} & $-0.411$\\
EfficientNet-B0 & CNN & 4.0  & 9.5   & 32  & 0.88 & \underline{\textbf{1.533}} & \underline{54.0} & \underline{0.144} & 1.668 & 49.7 & 0.049 & $+0.135$\\
Swin-T          & ViT & 27.5 & 65.0  & 42  & 1.15 & \underline{1.584} & \underline{\textbf{54.6}} & \underline{\textbf{0.223}} & 1.918 & 47.3 & 0.004 & $+0.334$\\
ResNet-50       & CNN & 23.5 & 55.6  & 334 & 9.15 & \underline{1.658} & \underline{49.9} & \underline{0.134} & 1.894 & 46.1 & $-0.019$ & $+0.236$\\
ViT-S/16        & ViT & 21.7 & 51.3  & 79  & 2.16 & 1.740 & 51.7 & 0.166 & \underline{1.660} & \underline{\textbf{53.4}} & \underline{\textbf{0.200}} & $-0.080$\\
ViT-B/16        & ViT & 85.8 & 202.8 & 201 & 5.51 & 1.790 & 49.4 & 0.210 & \underline{1.742} & \underline{52.5} & \underline{0.186} & $-0.048$\\
\midrule
Mean of six     & --  & 31.7 & 75.0  & 123 & 3.37 & 1.703 & 51.2 & 0.171 & 1.731 & 50.2 & 0.086 & $+0.028$\\
\midrule
 & & & & & & MAE\,$\downarrow$ & $\mathrm{Acc}_{\pm1}$\,$\uparrow$ & $\mathrm{Acc}_{\pm2}$\,$\uparrow$ & \multicolumn{3}{c}{QWK\,$\uparrow$ (s$\rightarrow$c / c$\rightarrow$s / mean)} & \\
\cmidrule(lr){7-9}\cmidrule(lr){10-12}
\textbf{InvarOrd (compressed, ours)} & Ordinal & \textbf{0.42} & 1.0 & 36.5 & 1.00 & \textbf{1.476} & \textbf{56.1} & 79.9 & \multicolumn{3}{c}{0.312 / 0.015 / 0.163} & --\\
\textbf{InvarOrd (same-cap., ours)}  & Ordinal & 1.07 & 2.5 & 90.3 & 2.47 & 1.486 & 54.7 & \textbf{82.0} & \multicolumn{3}{c}{0.171 / 0.053 / 0.112} & --\\
\bottomrule
\end{tabular}
\end{table*}

\begin{figure}[t]
\centering
\includegraphics[width=0.45\textwidth]{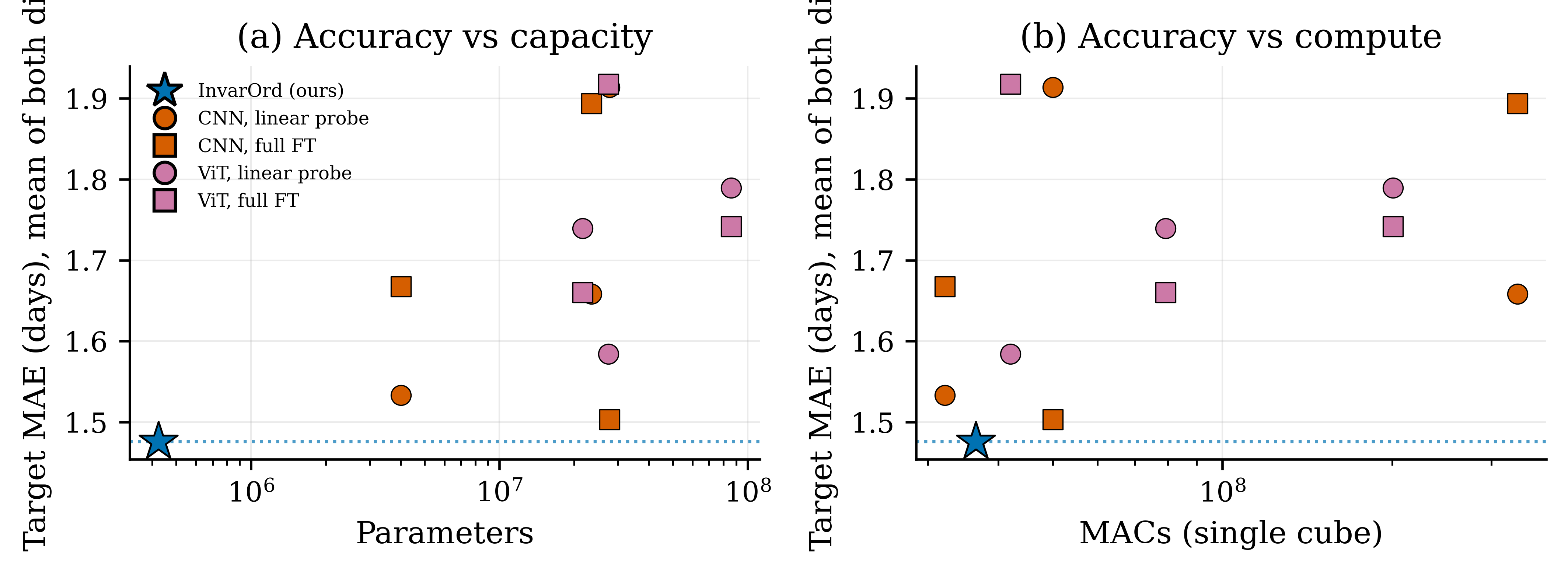}
\caption{Compressed student: MAE against (a) Parameters and (b) MACs.}
\label{fig:pareto}
\end{figure}

\begin{figure}[t]
\centering
\includegraphics[width=0.45\textwidth]{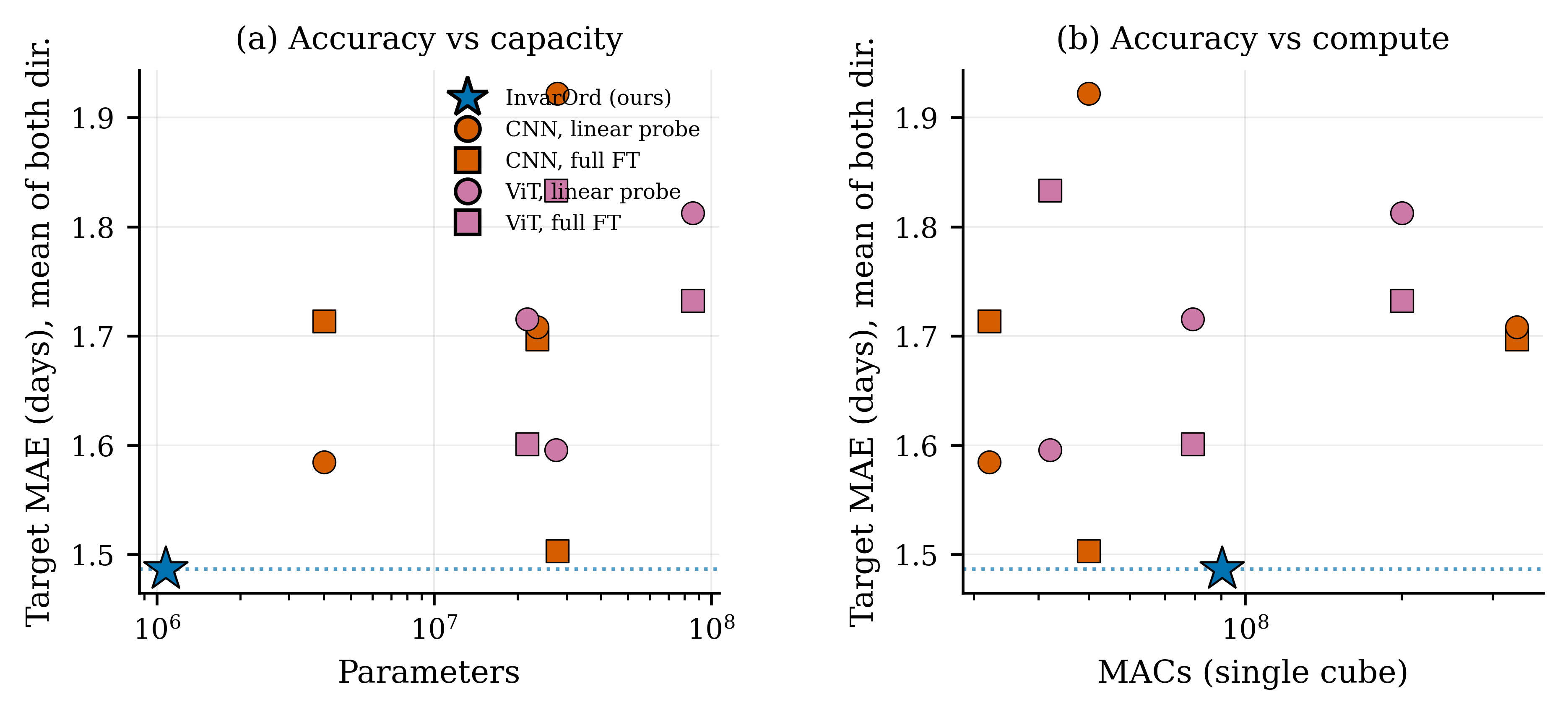}
\caption{Same-capacity student: MAE against (a) Parameters and (b) MACs.}
\label{fig:paretopaired}
\end{figure}

\begin{figure*}[t]
\centering
\includegraphics[width=0.80\textwidth,trim=0 0 0 0.32in,clip]{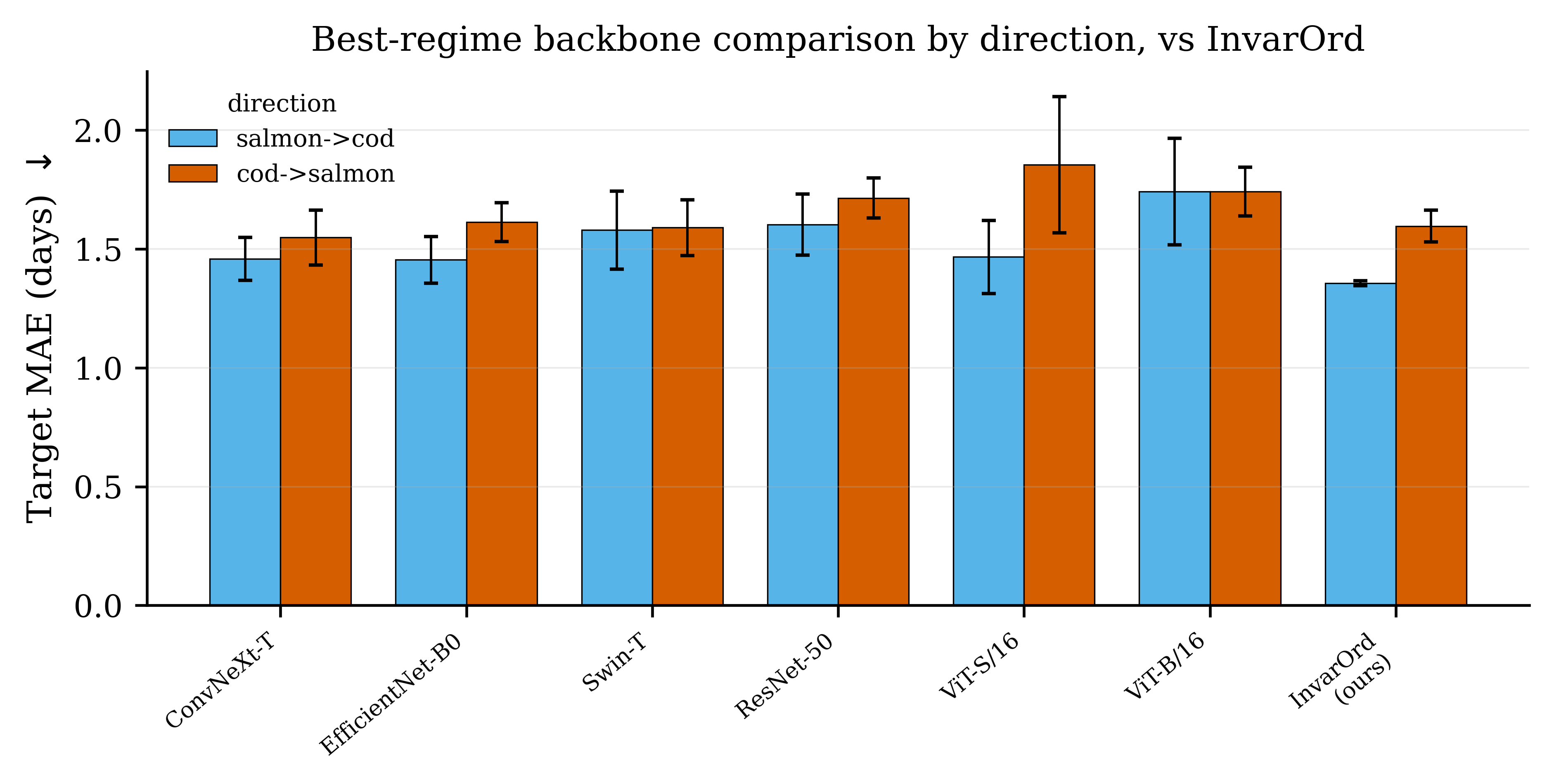}
\caption{Per-direction MAE of each backbone at its stronger regime;
InvarOrd rightmost. Error bars: SD over seeds.}
\label{fig:perbackbone}
\end{figure*}

Compute savings are smaller and uneven, $1.4\times$ against ConvNeXt-T and
up to $9.2\times$ against ResNet-50, and EfficientNet-B0 is $12\%$ cheaper in
MACs at a higher MAE. Fig.~\ref{fig:pareto} places every method on the
accuracy and cost planes, Fig.~\ref{fig:paretopaired} repeats the parameter
view for each student, and Fig.~\ref{fig:perbackbone} shows that the
ordering holds in each direction separately. InvarOrd does not lead on every
measure: Swin-T and both ViTs exceed it on QWK by $0.023$ to $0.060$, at $0.11$
to $0.27$ days higher MAE. Ordinal agreement and absolute error are therefore
not interchangeable at this budget.

\begin{table}[t]
\centering
\caption{Each backbone at its stronger regime relative to the compressed
student. Positive $\Delta$ favours InvarOrd.}
\label{tab:tradeoff}
\footnotesize
\setlength{\tabcolsep}{3pt}
\begin{tabular}{@{}llrrrrr@{}}
\toprule
Backbone & Regime & $\times$Params & $\times$MACs & $\Delta$MAE & $\Delta\mathrm{Acc}_{\pm1}$ & $\Delta$QWK \\
\midrule
ConvNeXt-T      & FT & 65.7  & 1.37 & $+0.027$ & $+3.6$ & $+0.069$\\
EfficientNet-B0 & LP & 9.5   & 0.88 & $+0.057$ & $+2.1$ & $+0.019$\\
Swin-T          & LP & 65.0  & 1.15 & $+0.108$ & $+1.5$ & $-0.060$\\
ResNet-50       & LP & 55.6  & 9.15 & $+0.182$ & $+6.2$ & $+0.029$\\
ViT-S/16        & FT & 51.3  & 2.16 & $+0.184$ & $+2.7$ & $-0.037$\\
ViT-B/16        & FT & 202.8 & 5.51 & $+0.266$ & $+3.6$ & $-0.023$\\
\bottomrule
\end{tabular}
\end{table}

\begin{figure}[t]
\centering
\includegraphics[width=\columnwidth]{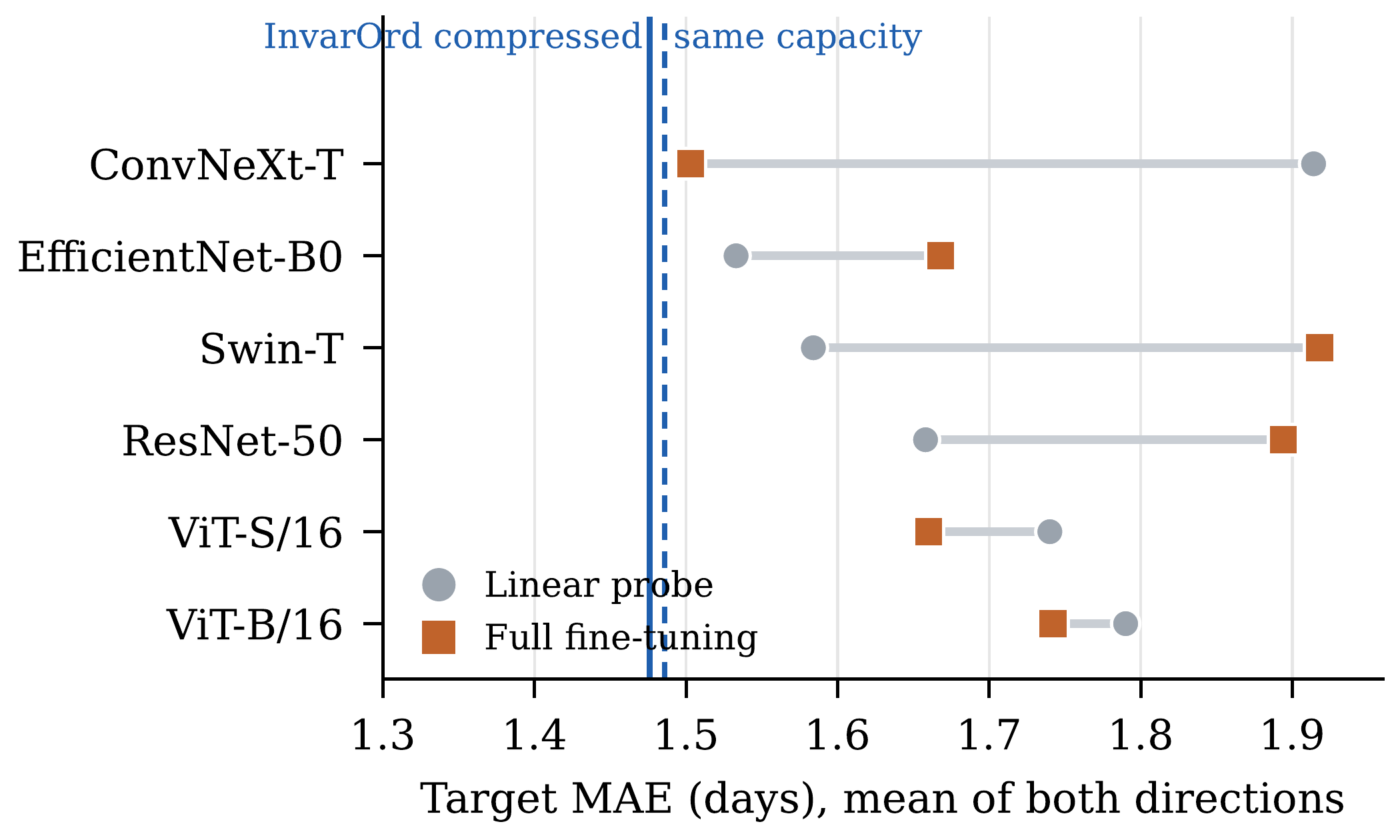}
\caption[LP and FT MAE per backbone]{LP (circle) and FT (square) MAE per backbone; vertical lines: the two InvarOrd students.}
\label{fig:regime}
\end{figure}

The adaptation regime matters as much as the backbone
(Table~\ref{tab:regime}, Fig.~\ref{fig:regime}). FT helps ConvNeXt-T by
$0.41$ days and both ViTs slightly, but hurts Swin-T, ResNet-50 and
EfficientNet-B0 by $0.14$ to $0.33$ days, so each regime wins for exactly
three of the six backbones. Averaged over the six, FT leaves MAE almost
unchanged ($1.731$ against $1.703$) but halves QWK ($0.086$ against $0.171$),
and three fine-tuned models fall to QWK $\le0.05$. This is consistent with
overfitting when millions of weights are tuned on $30$ cubes. No single
regime choice would have let the backbones match InvarOrd on MAE, whereas the
compact student is trained end to end at a scale suited to the label
budget.

\begin{table}[t]
\centering
\caption{LP against FT over six backbones (means). Wins: backbones for which
the regime gives lower MAE.}
\label{tab:regime}
\footnotesize
\begin{tabular}{@{}lccccc@{}}
\toprule
Regime & Mean MAE & Best MAE & $\mathrm{Acc}_{\pm1}$ & QWK & Wins \\
\midrule
Linear probe     & 1.703 & 1.533 & 51.2 & \textbf{0.171} & 3 / 6\\
Full fine-tuning & 1.731 & 1.503 & 50.2 & 0.086 & 3 / 6\\
\midrule
InvarOrd (compr.) & \textbf{1.476} & \textbf{1.476} & \textbf{56.1} & 0.163 & --\\
\bottomrule
\end{tabular}
\end{table}

\vspace{-10pt}
\subsection{Compression and Deployment (Q3)}
\label{subsec:students}

Distilling into the $423$K student costs nothing measurable. It matches or
beats the same-capacity student on mean MAE, $\mathrm{Acc}_{\pm1}$ and mean
QWK, and trails only on $\mathrm{Acc}_{\pm2}$ ($79.9$ against $82.0\%$)
(Table~\ref{tab:students}). For both students cod$\rightarrow$salmon is the
harder direction, mirroring the zero-shot asymmetry. MAE is stable across
seeds in both directions (SD $0.011$ and $0.067$ days), but
cod$\rightarrow$salmon QWK is close to zero and variable ($0.015\pm0.127$),
so ordinal agreement in that direction remains the open problem.

The compressed student runs at a measured $1.44$\,ms per cube, and Stages~2 and~3
re-adapt it to a new species in about $5$\,s from $30$ cubes
(Table~\ref{tab:deploy}), which makes on-instrument re-calibration
practical. Latency is measured per cube with batch size one on a single NVIDIA GPU, excluding data loading, and adaptation time is the wall-clock time of Stages~2 and~3 together. The latency gap to the larger student is small ($1.44$ against
$1.46$\,ms), most likely because fixed per-layer overheads dominate at this
scale, so the $2.54\times$ saving matters chiefly for memory and embedded
targets (Section~\ref{subsec:deploybound}).

\begin{table}[t]
\centering
\caption{Compressed vs. same-capacity student, $k=5$; $\pm$: SD over seeds.}
\label{tab:students}
\small
\begin{tabular}{@{}lcc@{}}
\toprule
Metric & Compressed & Same capacity \\
\midrule
Mean MAE\,$\downarrow$ & \textbf{1.476} & 1.486\\
MAE, sal$\rightarrow$cod $\downarrow$ & $\mathbf{1.356}\pm0.011$ & 1.369\\
MAE, cod$\rightarrow$sal $\downarrow$ & $\mathbf{1.596}\pm0.067$ & 1.604\\
$\mathrm{Acc}_{\pm1}$ (\%) $\uparrow$ & \textbf{56.1} & 54.7\\
$\mathrm{Acc}_{\pm2}$ (\%) $\uparrow$ & 79.9 & \textbf{82.0}\\
QWK, sal$\rightarrow$cod $\uparrow$ & $\mathbf{0.312}\pm0.033$ & 0.171\\
QWK, cod$\rightarrow$sal $\uparrow$ & $0.015\pm0.127$ & \textbf{0.053}\\
Mean QWK\,$\uparrow$ & \textbf{0.163} & 0.112\\
\bottomrule
\end{tabular}
\end{table}

\begin{table}[t]
\centering
\caption[Capacity and deployment cost]{Capacity and deployment cost. Latency and wall time are measured on one NVIDIA GPU; weight sizes follow from the parameter counts, and 8-bit sizes are prospective.}
\label{tab:deploy}
\footnotesize
\begin{tabular}{@{}lccc@{}}
\toprule
Quantity & Teacher & Compressed & Same capacity \\
\midrule
Width, embedding $(w,d)$ & $(64,256)$ & $(40,160)$ & $(64,256)$\\
Parameters & 1075\,K & \textbf{423\,K} & 1075\,K\\
MACs per cube & 90.3\,M & \textbf{36.5\,M} & 90.3\,M\\
Weights, 32/8-bit (MB) & 4.30/1.08 & \textbf{1.69/0.42} & 4.30/1.08\\
Latency (batch 1) & 1.46\,ms & \textbf{1.44\,ms} & 1.46\,ms\\
Compression & $1.00\times$ & $\mathbf{2.54\times}$ & $1.00\times$\\
\midrule
Target labels to adapt & \multicolumn{3}{c}{$5\times6=30$ cubes}\\
Stage 2 and 3 wall time & \multicolumn{3}{c}{about $5$\,s}\\
\bottomrule
\end{tabular}
\end{table}

\begin{table*}[t]
\centering
\caption{Ablation, same-capacity configuration, $k=5$, $n=6$ pairs.
$\Delta$MAE: variant minus InvarOrd (positive favours InvarOrd); $p$:
Wilcoxon; $d_z$: paired effect size.}
\label{tab:ablation}
\footnotesize
\begin{tabular}{@{}lllcccccccccc@{}}
\toprule
Variant & Stage 2 alignment & Pooling & $\lambda_2$ & Teacher & SSL
& MAE\,$\downarrow$ & SD & QWK\,$\uparrow$ & $\rho$\,$\uparrow$ & $\Delta$MAE & $p$ & $d_z$ \\
\midrule
\textbf{InvarOrd} & CORAL (shrinkage) & per rank & 0.3 & \checkmark & \checkmark & 1.486 & 0.134 & 0.112 & 0.183 & -- & -- & --\\
\midrule
\multicolumn{13}{l}{\textit{Alignment objective replaced}}\\
Rank-pooled MMD   & MMD (RBF)        & per rank & 0.3 & \checkmark & \checkmark & 1.500 & 0.068 & 0.000 & 0.102 & $+0.014$ & 0.563 & 0.21\\
Global MMD        & MMD (RBF)        & global   & 0.3 & \checkmark & \checkmark & 1.500 & 0.068 & 0.000 & 0.049 & $+0.014$ & 0.563 & 0.21\\
Adversarial       & DANN             & global   & 0.3 & \checkmark & \checkmark & 1.667 & 0.081 & 0.041 & 0.068 & $+0.180$ & 0.188 & 0.96\\
\midrule
\multicolumn{13}{l}{\textit{Component removed}}\\
w/o Stage 2       & none             & --       & 0.3 & \checkmark & \checkmark & 1.476 & 0.190 & \textbf{0.193} & \textbf{0.247} & $-0.010$ & 0.719 & $-0.13$\\
w/o $\mathcal{L}_{\mathrm{inv}}$ & CORAL (shrinkage) & per rank & 0.0 & \checkmark & \checkmark & \textbf{1.460} & 0.128 & 0.123 & 0.232 & $-0.026$ & 0.188 & $-0.80$\\
w/o SSL (supervised init.) & CORAL (shrinkage) & per rank & 0.3 & \checkmark & -- & 1.608 & 0.157 & 0.151 & 0.179 & $+0.122$ & 0.094 & 0.99\\
Target-only scratch & none           & --       & 0.0 & --         & -- & 1.773 & 0.209 & 0.152 & 0.152 & $+0.286$ & \textbf{0.031} & 2.63\\
\bottomrule
\end{tabular}
\end{table*}

\subsection{Does Matched-Rank Alignment Preserve Order? (Q4)}
\label{subsec:ablation}

The ablation uses the same-capacity student, which inherits the aligned
encoder in full (Eq.~\eqref{eq:inherit}), and we read it on two axes
(Tables~\ref{tab:ablation} and~\ref{tab:signif}).

\emph{Absolute error.} InvarOrd improves on target-only training by $0.286$
days ($p=0.031$, $d_z=2.63$), on adversarial alignment by $0.180$ days
($d_z=0.96$) and on supervised initialisation by $0.122$ days ($d_z=0.99$),
so self-supervised pretraining and a non-adversarial alignment both earn their
place. The last two effects are large even though $p$ exceeds $0.05$, given
the $0.031$ floor. Global MMD, no Stage~2 and no $\mathcal{L}_{\mathrm{inv}}$
lie within $0.026$ days, so we make no claim that matched-rank
alignment lowers MAE at $k=5$.

\begin{figure}[t]
\centering
\includegraphics[width=\columnwidth]{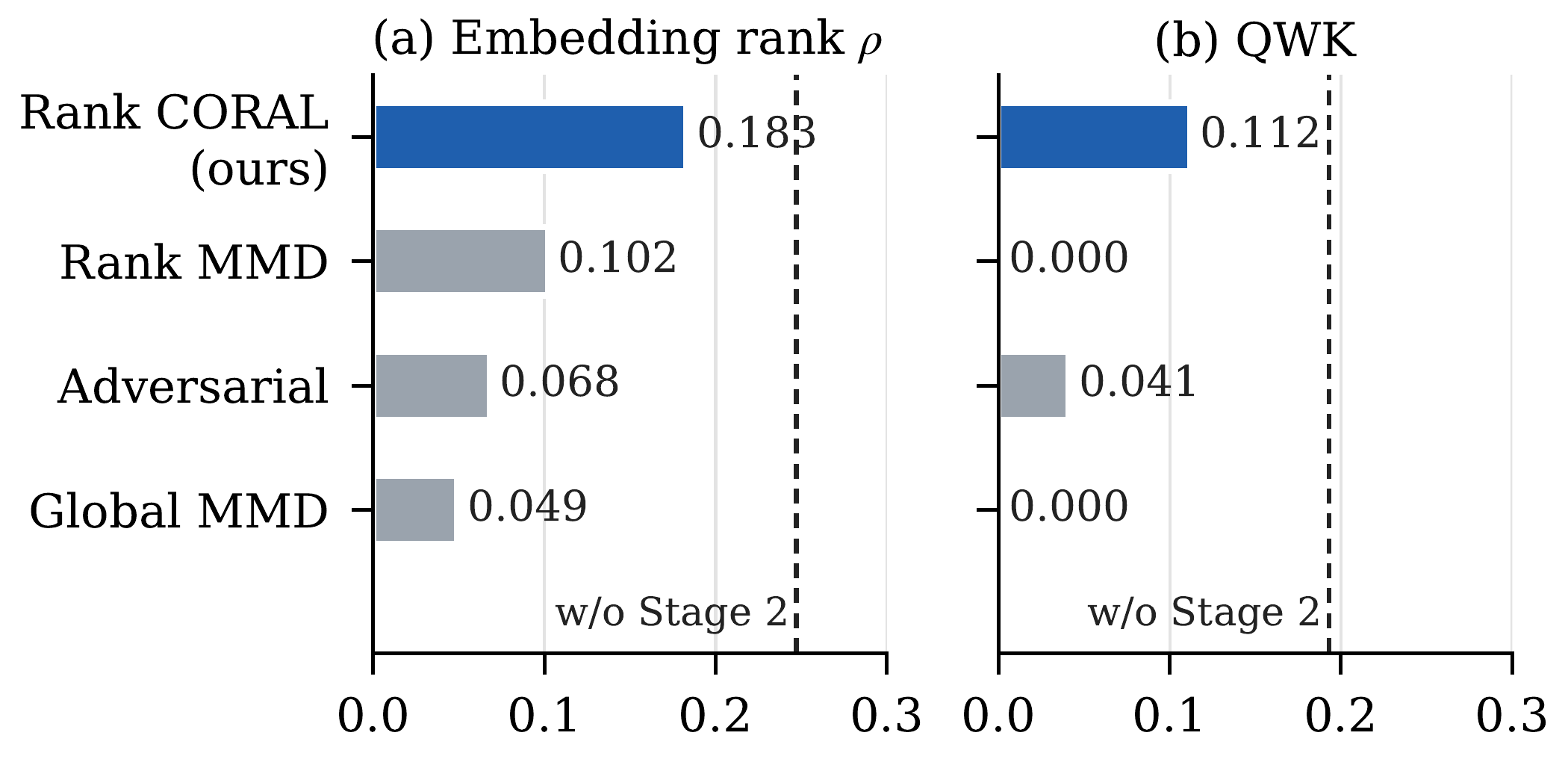}
\caption{(a) Embedding $\rho$ and (b) QWK per aligner; dashed: w/o Stage~2.}
\label{fig:ordalign}
\end{figure}

\begin{figure}[t]
\centering
\includegraphics[width=\columnwidth]{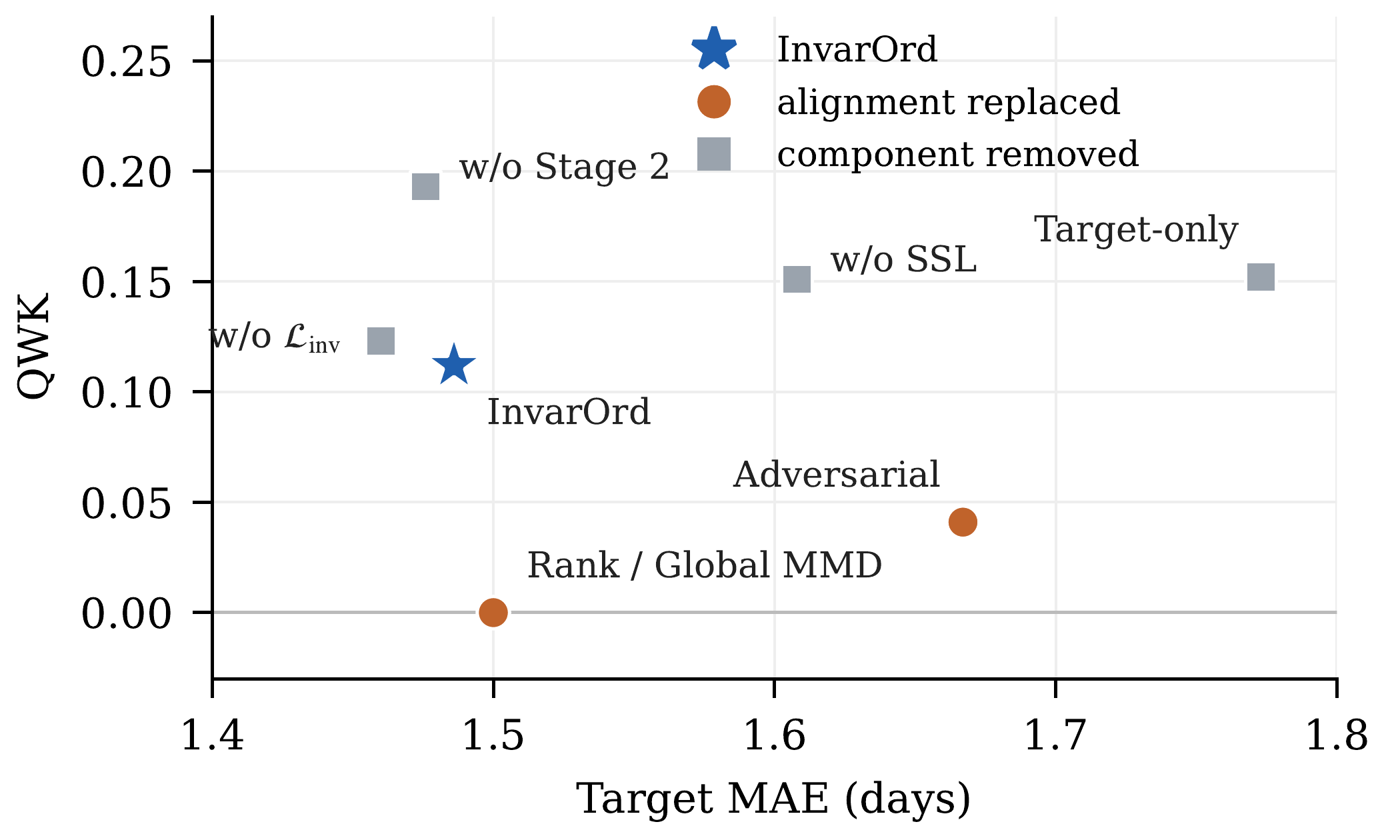}
\caption{Ablation variants on the MAE and QWK plane. Replacing alignment
(circles) lowers QWK; removing components (squares) raises MAE.}
\label{fig:ablmap}
\end{figure}

\emph{Ordinal structure.} This is where the aligners separate, consistent
with Proposition~\ref{prop:invariance} (Figs.~\ref{fig:ordalign}
and~\ref{fig:ablmap}). The two global objectives, which see the
between-rank scatter of Eq.~\eqref{eq:totalcov}, retain the least ordinal
structure: global MMD gives $\rho=0.049$ and adversarial alignment
$\rho=0.068$. Global MMD also reaches QWK $0.000$, so its predictions carry
no ordinal information even though its MAE is only $0.014$ days from ours;
MAE alone would hide this failure. Pooling the same MMD by rank doubles
$\rho$ to $0.102$ but leaves QWK at zero. Pooling by rank and matching
second-order statistics, the one objective shown to be invariant to rank
means, gives the highest $\rho$ ($0.183$) and QWK ($0.112$) of any aligner.
Read as controlled contrasts, pooling MMD by rank (statistic fixed) raises $\rho$ from $0.049$ to $0.102$ but leaves QWK at zero, while switching to the mean-invariant covariance (pooling fixed) adds $0.081$ to $\rho$ and lifts QWK to $0.112$ at no cost in MAE, the closest test our design allows of Proposition~\ref{prop:invariance}. By comparison, SSL and the full pipeline act mainly on accuracy (MAE $-0.122$ and $-0.286$ days, QWK about $0.04$ lower), and replacing adversarial alignment improves both.

\begin{figure*}[t]
\centering
\begin{minipage}[b]{0.49\textwidth}
\centering
\includegraphics[width=\linewidth]{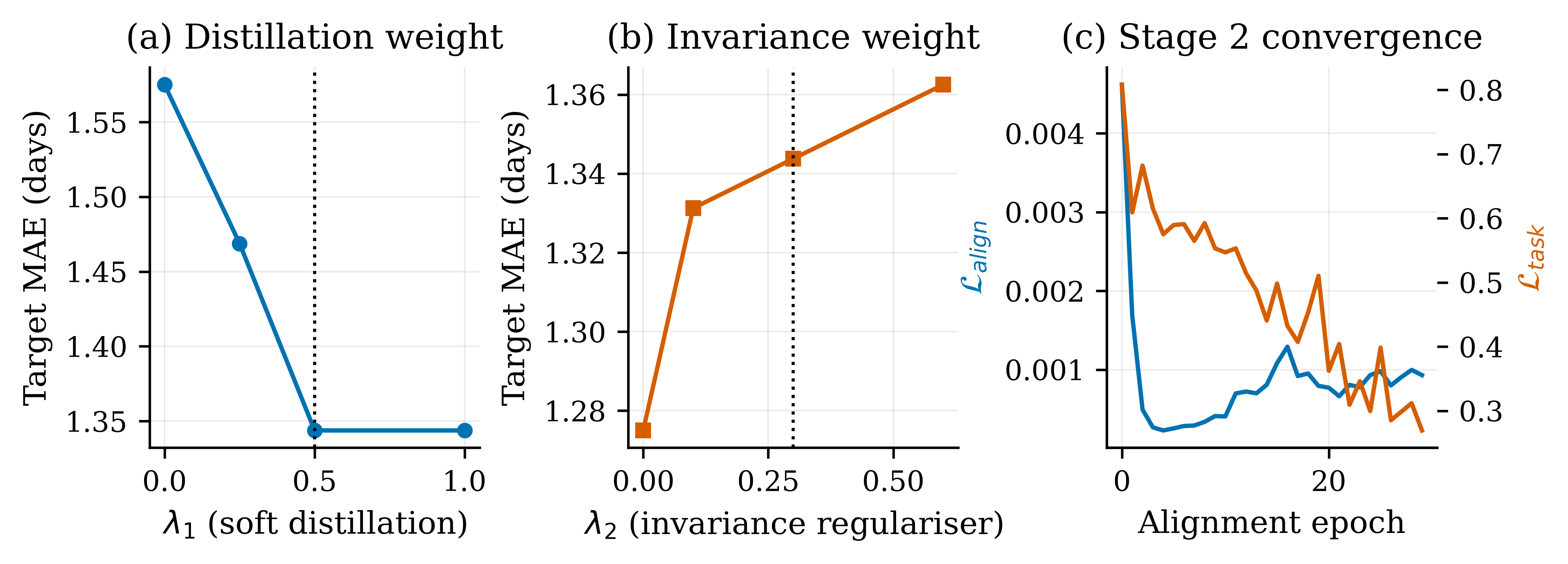}\\[3pt]
\includegraphics[width=\linewidth]{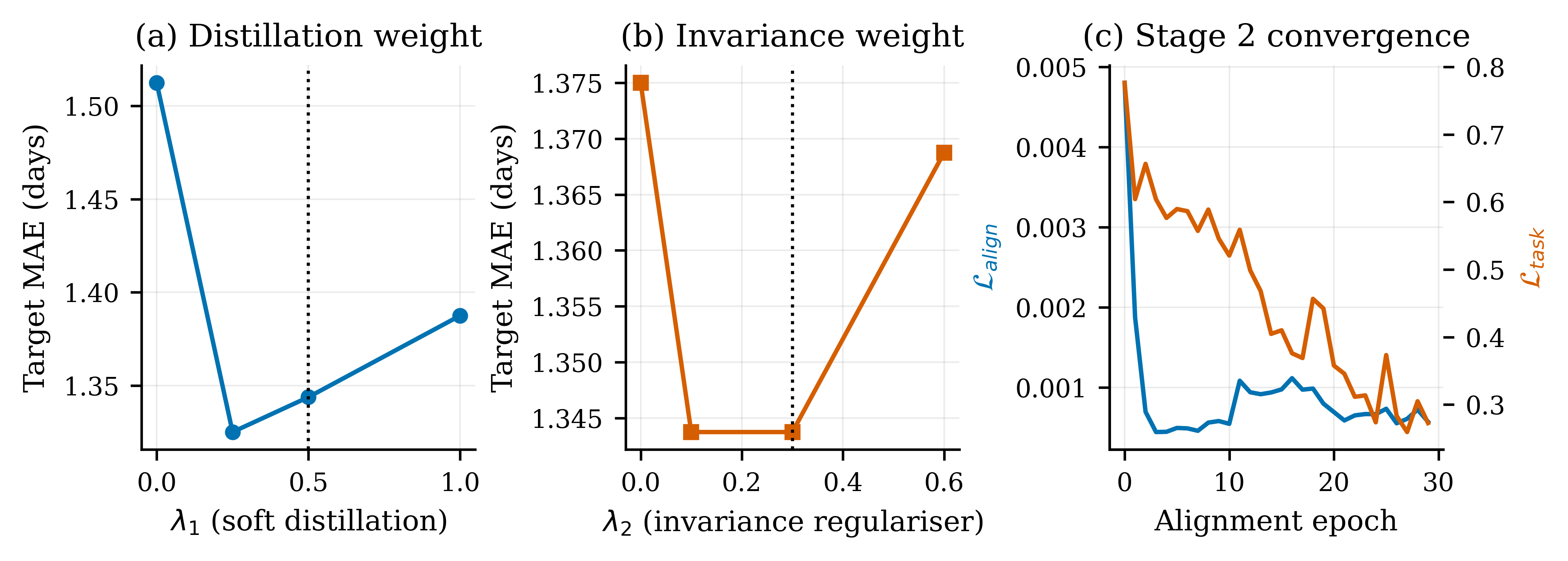}
\caption{Sensitivity to (a) the distillation weight $\lambda_1$ and (b) the
invariance weight $\lambda_2$, with defaults dotted, and (c) Stage-2 alignment
and task losses. Top: compressed; bottom: same capacity.}
\label{fig:sensitivity}
\end{minipage}\hfill
\begin{minipage}[b]{0.49\textwidth}
\centering
\includegraphics[width=\linewidth]{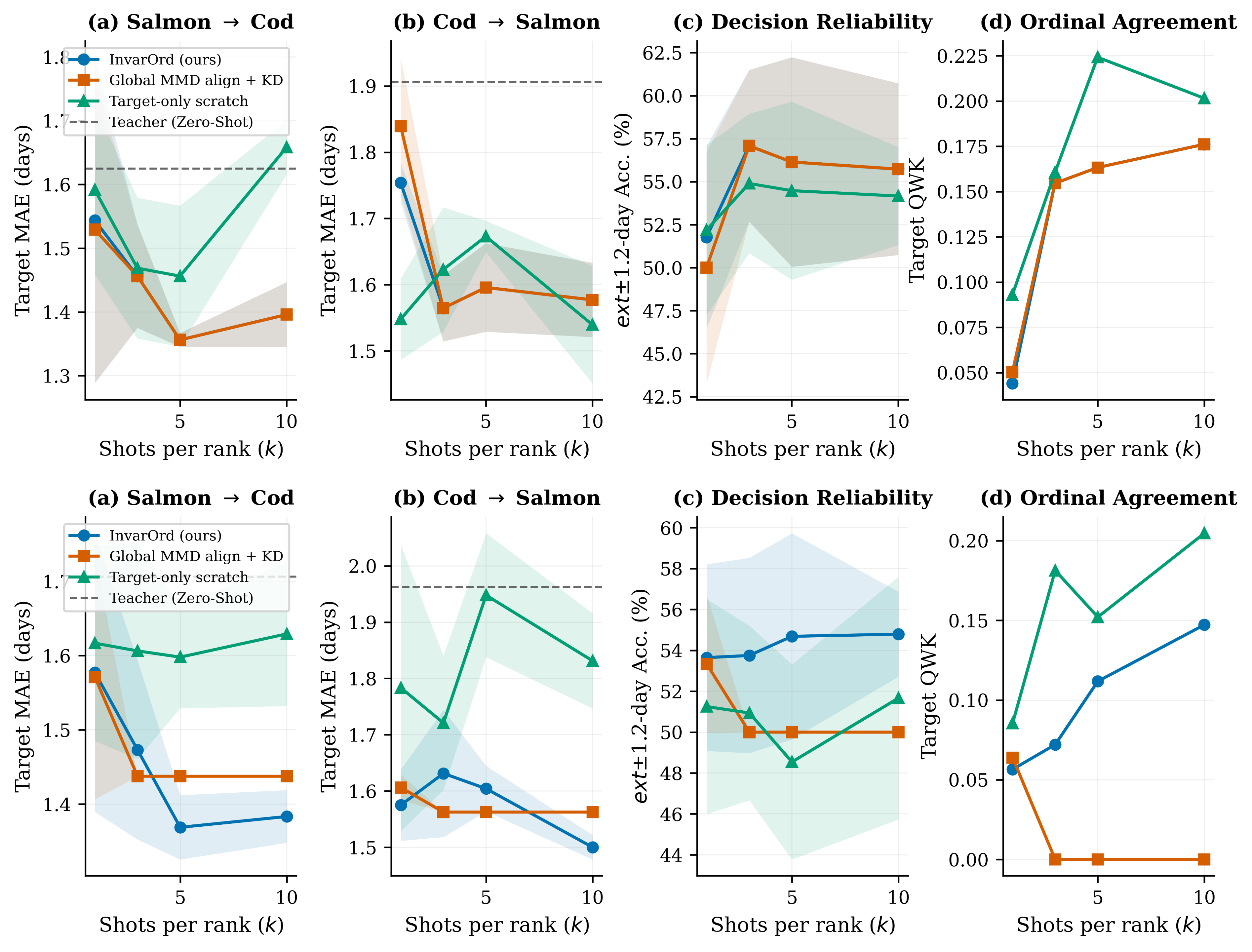}
\caption[Shot sweep]{Shot sweep over $k$, both
directions: (a,b) MAE, (c) $\mathrm{Acc}_{\pm1}$, (d) QWK. Top: compressed
student, where InvarOrd and global MMD overlap; bottom: same-capacity student, where InvarOrd leads on QWK.}
\label{fig:shotsweep}
\end{minipage}
\end{figure*}

Two variants improve on the full method, in line with Section~\ref{subsec:align}. Removing $\mathcal{L}_{\mathrm{inv}}$ lowers MAE by $0.026$ days and raises $\rho$ to $0.232$: as a global MMD it sees the between-rank scatter of Eq.~\eqref{eq:totalcov}, so it can pull against $\mathcal{L}^{\mathrm{rank}}_{\mathrm{align}}$. The effect is small and inconsistent in sign: in the sweep of Fig.~\ref{fig:sensitivity}, $\lambda_2=0$ is best for the compressed student but worst for the same-capacity one, whereas Table~\ref{tab:ablation}, averaged over both directions and all seeds, favours it for the latter, so we keep $\lambda_2=0.3$ as a single setting.

Without Stage~2, $\rho$ and QWK are higher still ($0.247$ and $0.193$). At $k=5$ the $30$ anchors already fix the order through $\mathcal{L}_{\mathrm{ord}}$, so any alignment term trades some ordinal sharpness for invariance; alignment should matter more at smaller $k$ or larger species gaps. The contribution of matched-rank alignment is therefore precise
and bounded. When alignment is used, it keeps the most ordinal structure,
whereas global MMD erases it; it does not lower MAE at $k=5$, and it reaches the deployed model only through an inherited encoder (Section~\ref{subsec:compressmech}).

\begin{table}[t]
\centering
\caption{Paired tests against InvarOrd sorted by $d_z$ (bands: $0.2$ small,
$0.5$ medium, $0.8$ large).}
\label{tab:signif}
\footnotesize
\begin{tabular}{@{}lcccl@{}}
\toprule
InvarOrd vs. & $\Delta$MAE & $p$ & $d_z$ & Reading \\
\midrule
Target-only scratch & $+0.286$ & \textbf{0.031} & 2.63 & large, significant\\
w/o SSL             & $+0.122$ & 0.094 & 0.99 & large, at power limit\\
Adversarial         & $+0.180$ & 0.188 & 0.96 & large, at power limit\\
Rank-pooled MMD     & $+0.014$ & 0.563 & 0.21 & small, none\\
Global MMD          & $+0.014$ & 0.563 & 0.21 & small, none\\
w/o Stage 2         & $-0.010$ & 0.719 & $-0.13$ & negligible\\
w/o $\mathcal{L}_{\mathrm{inv}}$ & $-0.026$ & 0.188 & $-0.80$ & large, favours variant\\
\bottomrule
\end{tabular}
\end{table}

\subsection{Label Efficiency (Q5) and Limitations}
\label{subsec:limits}

Over $k\in\{1,3,5,10\}$, MAE falls steeply from $k=1$ and levels off
between $k=3$ and $5$ in both directions (Fig.~\ref{fig:shotsweep}), so the
operating point $k=5$ sits on the plateau rather than at a favourable peak.
The two students differ with respect to alignment. For the compressed
student the InvarOrd and global MMD curves overlap at every $k$, while for
the same-capacity student, which inherits the aligned encoder, InvarOrd
separates from global MMD, most visibly on QWK. This agrees with the ordinal
analysis above.

\section{\texorpdfstring{Discussion}{Discussion}}
\label{sec:discussion}

\subsection{\texorpdfstring{Why Rank-Invariant Alignment Preserves Order}{Why Rank-Invariant Alignment Preserves Order}}
\label{subsec:compressmech}
The central strength of InvarOrd is that its alignment term gives no incentive to trade freshness order for domain agreement, and the evidence for this is both theoretical and empirical. Proposition~\ref{prop:invariance} shows that the per-rank covariance discrepancy is blind to rank means, so minimising it cannot shrink the between-rank scatter of Eq.~\eqref{eq:totalcov}. The ablation is consistent with this: at the same MAE, global MMD collapses QWK to zero, whereas rank-invariant alignment keeps the highest $\rho$ and QWK of all aligners (Table~\ref{tab:ablation}). The analysis also explains how this advantage reaches the deployed model. By Eq.~\eqref{eq:kdgrad}, output-level distillation transfers the teacher's ordinal score along $\mathbf u_{\mathrm s}$, and it does so efficiently: the $423$K student matches the accuracy of the same-capacity student at $2.54\times$ compression. The full within-rank geometry is carried by the inherited encoder, which is why the same-capacity student retains the ordinal gain in full (Fig.~\ref{fig:shotsweep}). InvarOrd therefore offers two well-understood operating points: maximum compression with no loss in accuracy, or full ordinal preservation at teacher size, and relational distillation \cite{tung2019spkd} offers a route to combine them.
\vspace{-5pt}

\subsection{\texorpdfstring{Stable Multi-Objective Training}{Stable Multi-Objective Training}}
\label{subsec:lossbal}
Although Stage~3 combines five terms, its balance comes from the design rather than from fine tuning. Each term is normalised so that its weight keeps its meaning (Section~\ref{subsec:kd}), and the warm-up of Eq.~\eqref{eq:warmup} lets the ordinal and distillation terms establish the ranks first. By Eq.~\eqref{eq:kdgrad}, the supervised terms act only along $\mathbf u_{\mathrm s}$ while the alignment terms shape the remaining directions, so the two groups do not compete. Accordingly, MAE varies by at most about $0.2$ days over $\lambda_1\in[0,1]$ and by under $0.07$ days over $\lambda_2\in[0,0.6]$ (Fig.~\ref{fig:sensitivity}). The same principles carry over to a new species pair: keep each auxiliary term's gradient-norm ratio to $\mathcal{L}_{\mathrm{ord}}$, as GradNorm \cite{chen2018gradnorm} and uncertainty weighting \cite{kendall2018multitask} automate, rather than its raw weight.
\vspace{-5pt}

\subsection{\texorpdfstring{Robustness Across Species and Tissue}{Robustness Across Species and Tissue}}
\label{subsec:species}
InvarOrd contains no species-specific component, because alignment is defined by freshness ranks, not by species, and it is tested on a deliberately demanding pair. Salmon is fatty and coloured mainly by astaxanthin \cite{folkestad2008salmon}; cod is lean and white, and in both species myoglobin is confined largely to the dark lateral muscle. Their spoilage also differs: lean gadoids degrade mainly through trimethylamine oxide and non-protein nitrogen \cite{gram1996spoilage}, while lipid oxidation and pigment loss add visible-range changes in a fatty fish. Freshness therefore moves the two spectra in partly different ways (Fig.~\ref{fig:spectra}), which is why zero-shot transfer fails and why cod$\rightarrow$salmon is the harder direction. From only $30$ cubes, InvarOrd nevertheless lowers target MAE well below the zero-shot level in both directions (Tables~\ref{tab:teacher} and~\ref{tab:students}). Within a fillet, dorsal muscle, belly flap, myosepta and dark muscle differ in fat and scattering. Spatial pooling over each $128\times128$-pixel region and 3-day bins reduce sensitivity to this variation, and on a sorting line a fillet can be graded by aggregating the cumulative probabilities of several cubes.
\vspace{-5pt}

\subsection{\texorpdfstring{Deployment Readiness}{Deployment Readiness}}
\label{subsec:deploybound}
The student is small enough for embedded use: its $1.69\,$MB of 32-bit weights, $36.5\,$M MACs and $262\,$KB per input cube follow from the architecture, whereas the $0.42\,$MB 8-bit size is a prospective estimate, since post-training quantisation \cite{jacob2018quant,nagel2021quant} has not yet been applied. The $1.44\,$ms per cube was measured on an NVIDIA GPU; at this rate inference is likely to outpace push-broom acquisition, but embedded latency remains to be measured. Static per-channel quantisation with similarity-preserving calibration \cite{padeiro2026spq} is a natural path for a model whose embedding geometry carries the ordinal information. The decisive practical advantage is label efficiency. Re-adaptation from $30$ labelled cubes in about $5\,$s on the same GPU means that drift from lamp ageing, window fouling or a new supplier can be corrected on site with a small reference set, a routine step rather than a full re-collection campaign.
\vspace{-5pt}

\subsection{\texorpdfstring{Limitations and Future Work}{Limitations and Future Work}}
\label{subsec:limitations}
Three seeds and two directions bound the resolution of the paired tests to MAE differences of about $0.05$ days; one species pair and storage-time labels are evaluated; and the ordinal gain of alignment reaches the compressed student only through its distilled score.

We will extend InvarOrd to further species with paired TVB-N and microbial references, add geometry-preserving distillation \cite{tung2019spkd} to carry the full ordinal structure into compact students, and quantise the student and measure its latency on embedded hardware.
\vspace{-5pt}

\section{Conclusion}
\label{sec:conclusion}

This article presented InvarOrd, which combines self-supervised spectral pretraining,
rank-invariant shrinkage-CORAL alignment and ordinal knowledge
distillation for label-efficient cross-species hyperspectral fish freshness assessment. Experiments on salmon and cod demonstrate competitive
freshness prediction with a compact student, achieving a mean MAE of
$1.476$ days and $\mathrm{Acc}_{\pm1}$ of $56.1\%$ with $2.54\times$
fewer parameters than the teacher. Its performance is comparable to
that of substantially larger ImageNet-pretrained backbones, while rank-invariant alignment demonstrates an advantage in preserving
ordinal structure over the evaluated alignment alternatives in the
same-capacity setting. These findings support the potential of
InvarOrd for resource-efficient freshness assessment while identifying
the preservation of ordinal information during compression as an
important direction for future research beyond the agri-food industry.

\section*{Acknowledgments}
The authors thank the School of BEEC at Leeds Beckett University for computational resources and logistic support.

\section*{Data and Codes}
The HFQA data are available at Zenodo \cite{alam2026hfqa}; The code will be made publicly available for reproducibility. 


\vfill

\end{document}